\documentclass{article}

\usepackage[preprint]{corl_2026} 
\usepackage{times}

\usepackage[numbers]{natbib}
\usepackage{multicol}
\usepackage{amsmath}
\usepackage{amssymb}
\usepackage{graphicx}
\usepackage{bm}
\usepackage{caption}
\usepackage{booktabs}   
\usepackage{multirow}   
\usepackage{float}
\usepackage{needspace}

\definecolor{linkblue}{rgb}{0,0.4,0.8}

\hypersetup{
    colorlinks=true,
    urlcolor=linkblue,
    pdftitle={Play2Perfect: What Matters in Dexterous Play Pretraining for Precise Assembly?},
    pdfauthor={Tyler Ga Wei Lum, Kushal Kedia, C. Karen Liu, Jeannette Bohg},
    pdfsubject={Dexterous play pretraining for precise robotic assembly},
    pdfkeywords={reinforcement learning; dexterous manipulation; sim-to-real; robotic assembly}
}
\usepackage{enumitem}

\usepackage{makecell}

\author{
    Tyler Ga Wei Lum$^{*1}$ \quad
    Kushal Kedia$^{*2}$ \quad
    C.~Karen Liu$^{\dagger 1}$ \quad
    Jeannette Bohg$^{\dagger 1}$ \\[0.6em]
    $^1$Stanford University \quad
    $^2$Cornell University \quad
    $^{*}$Equal contribution \quad
    $^{\dagger}$Equal advising \\[0.8em]
    \href{https://play2perfect.github.io}
    {\textbf{\textcolor{linkblue}{play2perfect.github.io}}}
}

\newcommand{\methodname}{{\textit{Play2Perfect}}}

\title{\methodname: What Matters in Dexterous Play Pretraining for Precise Assembly?}

\begin{document}
{%
    \maketitle
    \vspace{-7mm}  
    \begin{center}
      \captionsetup{type=figure}
      \includegraphics[width=0.98\textwidth]{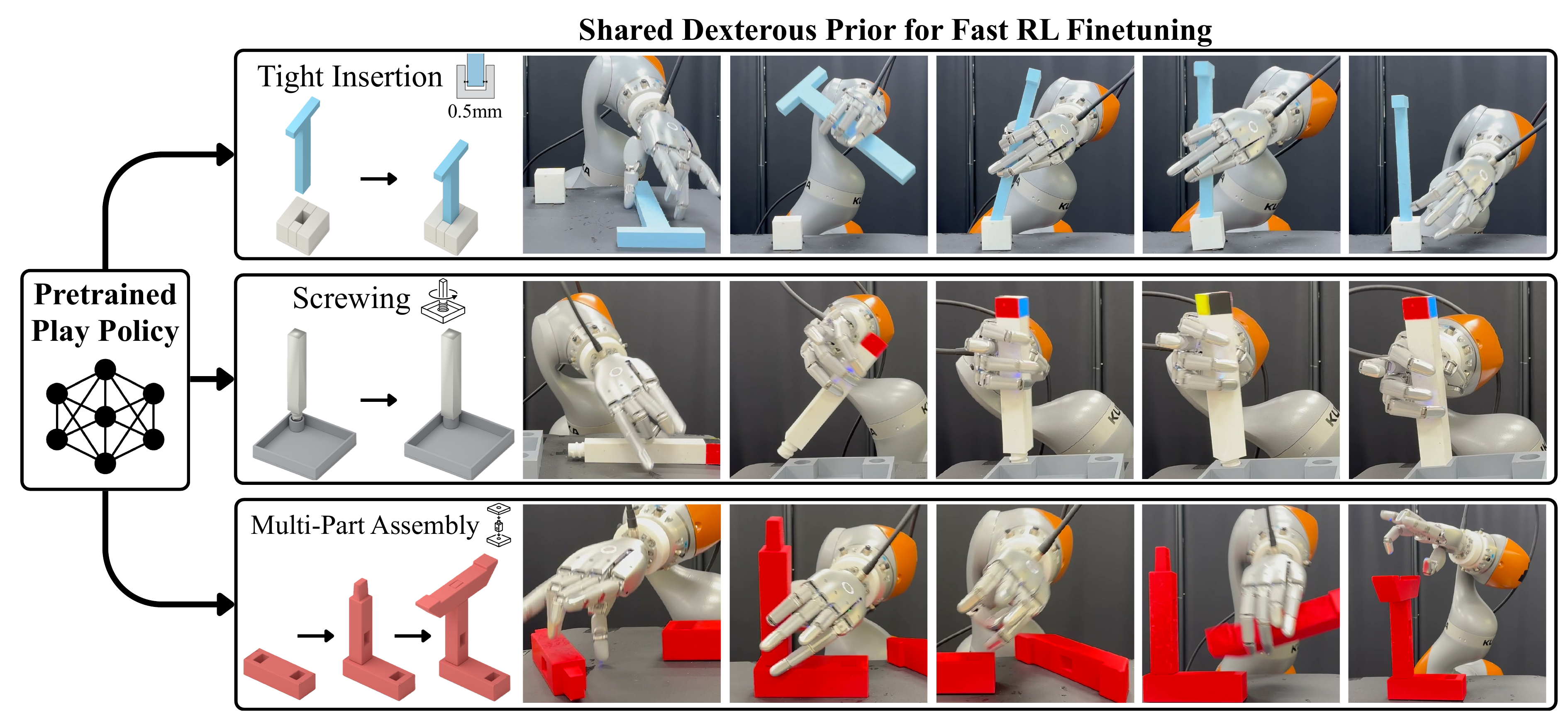}
      \captionof{figure}{
        \textbf{\methodname{} Overview.} Before a robot can perfect precise assembly, it first learns to play. We pretrain a single goal-conditioned RL policy on task-agnostic dexterous object manipulation, producing a reusable prior for grasping, in-hand reorientation, and 6D pose control. This pretrained play policy is then finetuned in sparse-reward RL environments derived from CAD designs to solve diverse contact-rich assembly tasks, including tight insertions, screwing, and multi-part assembly.
      }
      \label{fig:introduction}
    \end{center}
}

\thispagestyle{empty}
\pagestyle{empty}

\begin{abstract}
Multi-fingered robots promise the speed and dexterity of human hands, yet challenging problems such as precise assembly have remained out of reach. These tasks are contact-rich, making data collection for imitation learning difficult, and sparse-reward, making direct exploration with reinforcement learning (RL) intractable. Consequently, prior work has made progress by structuring the problem with specialized grippers, tool attachments, and environment fixtures. In this work, we argue that before a robot can \emph{perfect} precise assembly, it must first learn to \emph{play}. We further ask the question: what factors in the process of learning to \emph{play} matter for precise assembly? We propose \methodname, an RL framework for task-agnostic pretraining through \emph{play} on diverse objects and goals, which is then \emph{perfected} on precise assembly. The goal of \emph{play} is to acquire reusable manipulation priors, such as grasping, in-hand reorientation and pose reaching. Finetuning then adapts this general prior to assembly, focusing exploration on the final contact-rich, high-precision interactions needed for success. We systematically study key design choices in \emph{play} pretraining, including object diversity, training objective, trajectory diversity, and goal precision. We show that our prior is 33x more sample-efficient than RL training from scratch, even when provided with dense, multi-stage rewards. We demonstrate zero-shot sim-to-real transfer, achieving 60\% success on tight insertions with only 0.5 mm contact clearance, and over 50\% success on long-horizon multi-part assembly and screwing.
\end{abstract}


\section{Introduction}

Robots with multi-fingered hands hold the potential to bring speed and dexterity to the diverse tasks humans perform with their hands. Yet, this flexibility comes at the cost of controlling many degrees of freedom through contact, leaving challenging domains like precise assembly out of reach for current robot learning methods. On one hand, the contact-rich nature of assembly makes dexterous teleoperation challenging, so most imitation learning has focused on lower-precision pick-and-place tasks~\cite{wang2024dexcap, cheng2024open, bunny-visionpro, qin2023anyteleop, arunachalam2023dexterous}. On the other hand, assembly is sparse-reward, defined by a part's final pose, limiting the use of sim-to-real RL methods that require dense-reward shaping~\cite{wan2023unidexgrasp++, zhang2024dexgraspnet, chen2023visual, andrychowicz2020learning, handa2023dextreme, lin2024twisting, chen2023sequential}.

Prior work has made progress on assembly by adding structure to the problem. One approach is to modify the environment with custom fixtures that simplify grasping and insertion~\cite{luo2024fmb, shao2020learning}. Another is to modify the robot itself with specialized tool attachments or end-effectors that make the control problem easier~\cite{ha2020fit2form, shi2023robocook}. However, both strategies require per-assembly engineering of the hardware or environment. The use of robots with parallel-jaw grippers makes teleoperation feasible, enabling imitation learning~\cite{ankile2024juicer} and subsequent RL finetuning~\cite{ankile2025imitation}. Without teleoperation, RL methods often rely on dense, task-specific rewards~\cite{tang2023industreal, tang2024automate} or scripted multi-stage controllers~\cite{tian2025fabrica}. The reliance of these approaches on parallel-jaw grippers still limits their speed and dexterity.

Our goal is to tackle hard, sparse-reward problems such as precise assembly with dexterous hands, without relying on teleoperation. The core challenge is that the sparse reward defined by the final goal configuration of an assembly part offers limited training signal for RL. Starting from a random policy, the agent must discover grasping, in-hand reorientation, alignment, and contact-rich insertion before receiving any reward. Intuitively, before learning the hard problem of \emph{perfecting} precise assembly, a robot should first learn the easier problem of \emph{playing} with objects in free space. While the concept of learning from \emph{play} has been explored previously~\cite{lynch2020learning, wang2023mimicplay, kuang2026dex4d, kedia2026simtoolrealobjectcentricpolicyzeroshot}, it remains unclear what aspects of the \emph{play} pretraining recipe matter for downstream finetuning, especially for precise assembly. In this work, we systematically study the design choices that make \emph{play} useful for precise assembly, including object diversity, trajectory diversity, training objective, and goal precision. Across these studies, we find a consistent takeaway: \emph{play} pretraining transfers best when it forces the robot to learn in-hand manipulation using its fingers rather than movement with a fixed grasp.

We propose \methodname, a framework for dexterous pretraining on general objects and goals, followed by finetuning for precise assembly. In simulation, we pretrain a goal-conditioned \emph{play} policy via RL to manipulate diverse primitive objects to random target poses, inducing a task-agnostic manipulation prior. We then construct assembly finetuning environments from assembly benchmarks~\cite{tian2025fabrica, heo2023furniturebench}, using sparse rewards defined by the final part configuration. Across challenging assembly skills, this prior enables 33x more sample-efficient learning than RL from scratch, even when scratch policies are provided with dense, multi-stage rewards. We further demonstrate sim-to-real transfer, achieving 60\% success on tight insertions with 0.5 mm clearance and over 50\% success on long-horizon multi-part assembly and screwing. Our contributions are:

\begin{itemize}[leftmargin=*, itemsep=0pt, topsep=0pt, parsep=0pt, partopsep=0pt]
    \item A framework for precise assembly with dexterous hands that first learns a task-agnostic \emph{play} prior on general objects and goals, then \emph{perfects} it to new CAD-defined assembly tasks. 
    \item A systematic study of \emph{play} pretraining design choices that transfer to contact-rich, precise assembly, including object diversity, trajectory diversity, training objectives, and goal precision.
\end{itemize}
\section{Related Work}
\textbf{Manipulation with Multi-Fingered Robots.}
Prior work falls into two categories: imitation learning (IL) and reinforcement learning (RL). IL relies on collecting high-quality demonstrations, which can be obtained through human hand motion retargeting with motion-capture gloves~\cite{shaw2024bimanual,
wang2024dexcap}, VR devices~\cite{bunny-visionpro,
iyer2024open,lin2025learning, arunachalam2022holo}, camera inputs~\cite{qin2023anyteleop,
handa2019dexpilotvisionbasedteleoperation, sivakumar2022robotic}, or exoskeleton systems~\cite{tao2025dexwild, xu2025dexumi, fang2025dexop}. However, collecting demonstrations for contact-rich tasks remains difficult due to the embodiment gap between the human operator and the robot, as well as the lack of tactile feedback~\cite{chen2025dexforceextractingforceinformedactions,
Si-RSS-24,Human2RobotWholeBodyTransfer,pacchierotti2023cutaneous}. Sim-to-real RL offers a promising alternative, with recent progress on dexterous skills such as grasping~\cite{agarwal2023dexterousfunctionalgrasping, ye2025dex1b,
lum2024dextrahg, singh2025end, singh2024dextrah} and in-hand reorientation~\cite{chen2021system, chen2023visual,
liu2025dexndmclosingrealitygap}. Yet, these skills are largely performed in free space. Extending dexterous RL to contact-rich tasks relies on dense reward functions~\cite{lin2024twisting, chen2023sequential}, accurate human hand motion references~\cite{li2025maniptrans, lum2025crossinghumanrobotembodimentgap, mandi2025dexmachinafunctionalretargetingbimanual}, or warm-starting from teleoperation~\cite{si2025exostart, bauza2025demostart}. Closest to our work are methods that train task-agnostic \emph{play} controllers across diverse objects~\cite{yin2025dexteritygen, kedia2026simtoolrealobjectcentricpolicyzeroshot, kuang2026dex4d}. When combined with teleoperation~\cite{yin2025dexteritygen} or a human demonstration at test time~\cite{kedia2026simtoolrealobjectcentricpolicyzeroshot}, these controllers can generalize to unseen objects and tasks. However, they are deployed zero-shot and still struggle with precise, contact-rich tasks such as assembly. Instead, we view \emph{play} as dexterous pretraining: a general prior that is quickly specialized to precise assembly.

\textbf{Precise and Contact-Rich Assembly.}
Most progress in precise, contact-rich assembly has come from structuring the problem through task-specific hardware or environments. Prior work designs specialized gripper attachments or tools to simplify manipulation~\cite{ha2020fit2form, shi2023robocook}, or uses fixtures to reduce uncertainty in grasping, alignment, and insertion~\cite{shao2020learning, luo2024fmb}. While effective, these approaches require task-specific setup for each assembly problem. Learning-based assembly methods often rely on task-specific structure as well, including dense reward functions~\cite{tang2023industreal, tang2024automate}, scripted multi-stage controllers~\cite{tian2025fabrica}, or carefully designed curricula~\cite{jiang2024transic}. Other methods address the exploration problem with dense reset distributions~\cite{yin2026emergent} or teleoperation demonstrations~\cite{ankile2024juicer, ankile2025imitation}, but these still require task insight or task-specific data. In contrast, we pretrain a general dexterous \emph{play} prior without demos or knowledge of the downstream assembly task, and specialize it with sparse-reward RL.

\textbf{Pretraining for Dexterous Manipulation.} There is growing interest in pretraining broad dexterous priors to solve challenging manipulation tasks. For example, Vision Language Action (VLA) models~\cite{intelligence2025pi05, nvidia2025gr00t, barreiros2026careful} train on large datasets consisting of diverse tasks~\cite{o2024open, khazatsky2024droid}. However, these datasets are largely concentrated on parallel jaw gripper robots. Human videos on the internet are another promising option to acquire general priors for dexterous hands~\cite{zheng2026egoscale, yang2025egovla, qiu2025humanoid, tao2025dexwild}. However, human videos do not contain contact information, limiting their use to simple pick and place tasks and often requiring large amounts of in-domain robot finetuning data. Learning from \emph{play} is a promising approach to learn priors from task-agnostic data. Prior works still require teleoperation to collect robot trajectories or active human hand data collection~\cite{wang2023mimicplay, lynch2020learning}. Compared to these works, we train \emph{play} priors by learning to manipulate diverse objects via RL without requiring any demonstrations.

\begin{figure}[t!]
  \centering
  \includegraphics[width=\textwidth]{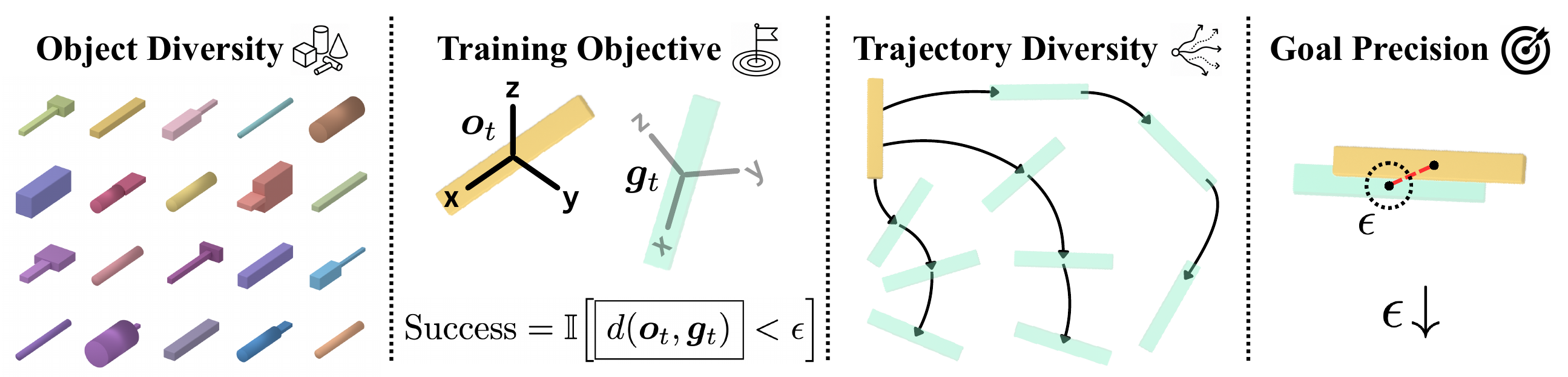}
    \caption{
    \textbf{What matters in dexterous \emph{play} pretraining?} We study the key factors that shape the learned manipulation prior. Our design emphasizes in-hand manipulation with fingers across diverse objects and trajectories, with 6D pose-reaching objectives and precise goal tolerances.
    }
  \label{fig:pretrain}
\end{figure}

\section{Play2Perfect}
\methodname\ is a framework for dexterous \emph{play} pretraining followed by sparse-reward RL finetuning for assembly. We first train a goal-conditioned policy in simulation to manipulate procedurally generated primitive objects to random poses in free-space. We then use the assembly CAD design to construct a sparse-reward RL environment for finetuning to learn precise, contact-rich goals.
\subsection{Dexterous Play Pretraining}
We first train a task-agnostic dexterous manipulation policy before specializing to any assembly task. The objective of \emph{play} is to acquire a reusable dexterous prior by manipulating diverse objects to random goal poses. Concretely, we formulate play as a goal-conditioned RL problem and train a policy $\pi_\theta(\bm{s}_t, \bm{o}_t, \bm{g}_t, \bm{\phi})$, where $\bm{s}_t$ denotes robot proprioception, $\bm{o}_t, \bm{g}_t \in SE(3)$ are the current and target object poses, and $\bm{\phi}$ encodes object geometry through its 3D bounding-box dimensions. A single policy controls both the arm and hand attached to it. Fig.~\ref{fig:pretrain} provides an overview of the key components of \emph{play} pretraining that enable transfer to downstream assembly tasks. More details of the environment design, observations, and action spaces are in Appendix~\ref{sec:play-pretraining-details}.

\textbf{Object Diversity.}
We aim to acquire broad dexterous competence across a wide range of objects, so that the resulting policy provides a useful initialization for downstream finetuning. To this end, we procedurally generate diverse primitive objects in simulation comprising cuboids and cylinders. Each object's dimensions are sampled from a broad distribution constrained to fit within the robot hand. We also randomize physical properties by varying object densities and attaching additional masses near object ends. This induces variation in center of mass and inertia, forcing the policy to learn object-control strategies that are not tied to a single geometry or mass distribution. We use primitive shapes to enable fast, stable simulation while keeping pretraining task-agnostic.

\textbf{Training Objective.}
Play pretraining should induce reusable dexterous skills for downstream assembly. Therefore, we train the robot to manipulate objects through a sequence of 6D goal poses: the first goal requires grasping and lifting from the table, while subsequent goals require in-hand control of object pose without drops. Each goal specifies both translation and rotation: translation teaches object motion across the workspace, while rotation encourages in-hand reorientation. The play reward comprises three terms: $r_{\mathrm{smooth}}$ regularizes actions for smooth control, $r_{\mathrm{grasp}}$ encourages lifting the object above the table, and $r_{\mathrm{goal}}$ rewards reaching the current 6D object goal. $r_{\mathrm{goal}}$ includes a large sparse success bonus when $d(\bm{o}_t,\bm{g}_t)<\epsilon$. By default, $d=d_{\rm pose}$ is a keypoint-based 6D pose distance that jointly captures translation and rotation error.

\textbf{Trajectory Diversity.}
We randomize the goal sequence in every play episode rather than training on fixed trajectories. The first goal is sampled broadly in the robot's workspace, while subsequent goals are sampled near the previous goal with significant rotations. This encourages learning of in-hand manipulation rather than simple arm movements with fixed grasps.

\textbf{Goal Precision.} The threshold $\epsilon$ controls the precision of the learned behavior, with $\epsilon=1\,\mathrm{cm}$ by default. A smaller threshold requires more accurate goal reaching, which is important for precise assembly, and forces fine object pose control via in-hand manipulation.

\begin{figure}[t!]
  \centering
  \includegraphics[width=0.85\textwidth]{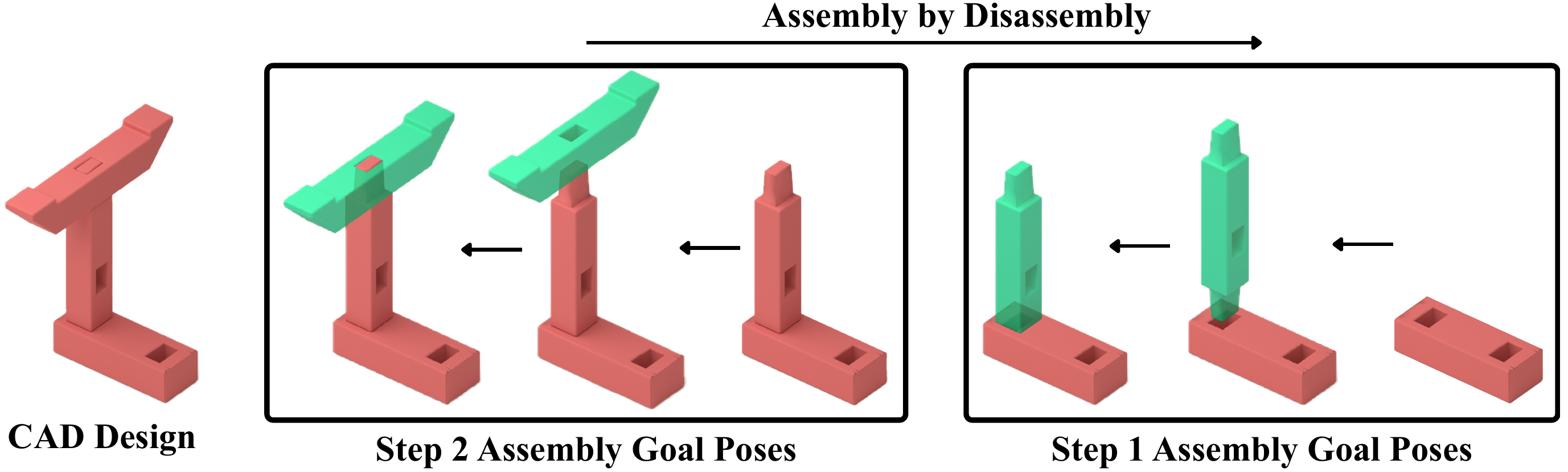}
  \caption{
    \textbf{Assembly-by-Disassembly.}
    Given a completed CAD assembly, we generate assembly steps by sequentially removing parts and reversing the disassembly sequence. Each step defines a sparse goal sequence: the final assembled pose and intermediate contact goals, e.g., pre-insert pose.
  }
  \label{fig:assembly}
\end{figure}


\subsection{RL Finetuning on Assembly Tasks}

After training a general \emph{play} policy for free-space object manipulation, we specialize it to specific assembly skills. Each assembly task is defined by the final desired part configuration and uses a sparse success reward. Finetuning adapts the pretrained manipulation prior to the contact-rich, high-precision interactions required for assembly. In this section, we describe how we construct simulation environments and derive sparse rewards directly from the assembly CAD design.

\textbf{Simulation Construction from CAD.}
Each assembly task is specified by a CAD design containing $K$ rigid parts $\mathcal{A}=\{p^i\}_{i=1}^K$ and their final assembled poses. We convert this design into a sequence of assembly steps using assembly-by-disassembly~\cite{tian2022assemble, yin2026emergent}: starting from the completed assembly, we identify feasible part removals and reverse this order to obtain an assembly sequence. Each step requires inserting a part $p^i$ into a fixture $f^i$, defined by the parts that have already been assembled. We instantiate each step as an RL environment with randomized part and fixture poses on the table.

\textbf{Inferring Sparse Assembly Rewards.}
Assembly finetuning uses only sparse success rewards derived from the CAD-specified part configurations. For each part-fixture pair, the CAD design specifies the desired relative transform $\bm{T}^{f^i}_{p^i}$ of the part $p^i$ in the fixture frame $f^i$. We denote the current part pose as $\bm{p}^i_t \in SE(3)$ and compute the final goal pose as $\bm{g}^i_m = \bm{f}^i_t \bm{T}^{f}_{p,m}$, where $\bm{T}^{f}_{p,m}$ is the CAD-derived part transform. For contact-rich assembly skills, we derive a small set of sparse contact goals $\mathcal{G}^i=\{\bm{g}^i_1,\ldots,\bm{g}^i_M\}$ by reversing the assembly motion, where the final goal $\bm{g}^i_M$ corresponds to the assembled pose (see Fig.~\ref{fig:assembly}). For insertion, this adds an aligned pre-insertion pose at the onset of contact. For screwing, we generate poses along the thread at fixed 90$^\circ$ rotational offsets. Environment construction, reward definitions, and goal progression are detailed in Appendix~\ref{sec:assembly-finetuning-details}.

\subsection{Training Details and Sim-to-Real Transfer}
\textbf{RL Algorithm and Domain Randomization.}
We train both play pretraining and assembly finetuning policies with Split and Aggregate Policy Gradients (SAPG)~\cite{sapg2024}, which prior work~\cite{kedia2026simtoolrealobjectcentricpolicyzeroshot} found to outperform PPO~\cite{PPO} for dexterous play. To enable sim-to-real transfer, we train all policies with domain randomization modeling action latency, proprioceptive observation delays, and noise in both current and goal object poses. RL algorithm details are provided in Appendix~\ref{sec:policy-rl-details}, and domain randomization settings for pretraining and finetuning are provided in Appendices~\ref{sec:play-pretraining-details} and~\ref{sec:assembly-finetuning-details}, respectively.

\textbf{CAD-Based Object Pose Tracking.}
At deployment, we reuse the assembly CAD meshes for real-world 6D pose tracking with FoundationPose~\cite{wen2024foundationpose}. We track both the current part pose and the fixture pose. The policy runs closed-loop at 60Hz, while the object pose tracking runs at 30Hz. The deployment pipeline is detailed in Appendix~\ref{app:inference}.

\vspace{-3mm}
\section{Experiments}
\label{sec:experiments}

We design experiments to answer the following questions:
\begin{enumerate}[leftmargin=*, label=\arabic*., nosep]  
    \item Can dense, task-specific rewards overcome the need for \emph{play} pretraining?
    \item Which \emph{play} pretraining design choices matter most for downstream assembly?
    \item Is RL finetuning necessary for precise, contact-rich assembly?
    \item Can \methodname{} policies transfer from simulation to real-world assembly tasks?
    \item Does \methodname{} transfer to everyday insertion tasks?
\end{enumerate}

\textbf{Task Design.} Our robot consists of a 22-DoF Sharpa five-fingered hand mounted on a 7-DoF KUKA iiwa 14 arm. We evaluate on three tasks: 1) \texttt{Tight-Insertion}, inserting a T-shaped peg into holes with increasingly tight contact clearances; 2) \texttt{Assemble-Beam}, a multi-part beam assembly constructed from Fabrica~\cite{tian2025fabrica}; and 3) \texttt{Screw-Leg}, screwing a furniture leg into a table fixture, constructed from FurnitureBench~\cite{heo2023furniturebench}. The original parts in Fabrica and FurnitureBench are small and designed for parallel-jaw grippers. We therefore 3D print 3$\times$-scale parts for \texttt{Assemble-Beam} and a 3$\times$-longer leg for \texttt{Screw-Leg}, making the tasks more suitable for dexterous hands and reliable for visual tracking under occlusion. All tasks are illustrated in Fig.~\ref{fig:introduction}.

\textbf{Evaluation Metrics.} We report two primary metrics: \textit{Success Rate}, which measures whether the assembly part reaches its final goal configuration within a tolerance of $\epsilon=1$cm, and \textit{Completion Time}, which measures the average time required to complete the full task, including approach, grasping, transport, and the final contact-rich interaction. In simulation, we report results over 500 rollouts with randomized initial part and fixture poses. In the real world, we evaluate each task over 10 or 20 rollouts using a fixed fixture pose and randomized initial part poses.

\subsection{Can Dense Rewards Overcome the Need for \emph{Play}?}
\begin{figure}[t!]
  \centering
  \includegraphics[width=\textwidth]{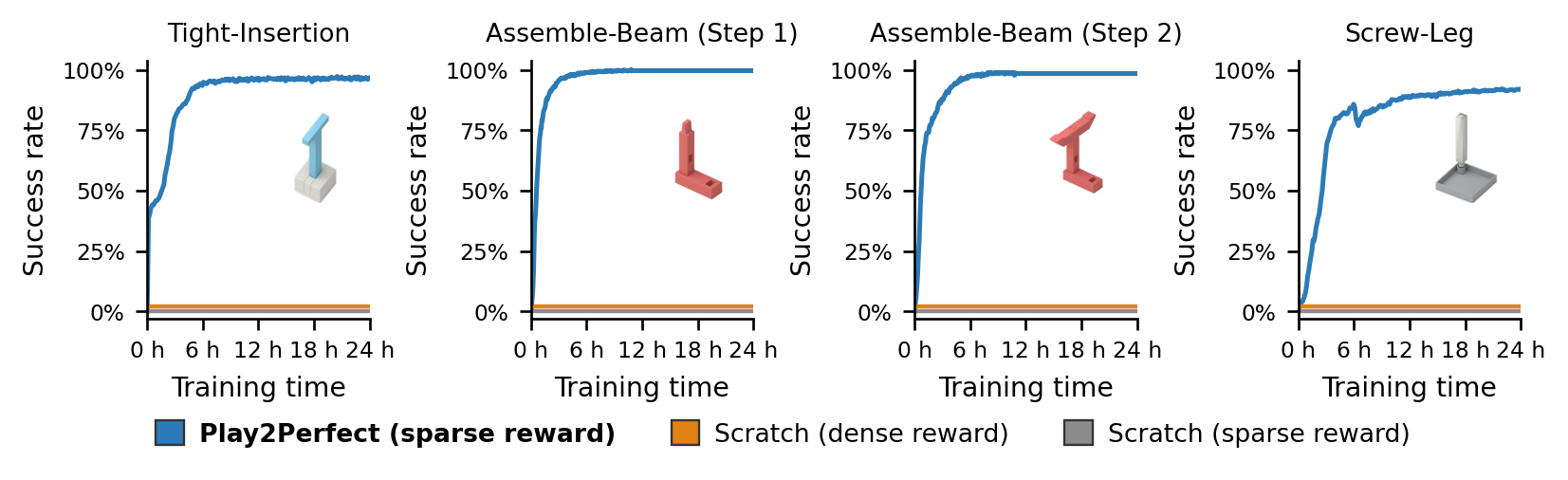}
  \caption{
     \textbf{Dexterous Play Pretraining Enables Efficient Downstream Assembly Learning.}   Across four contact-rich assembly tasks, \methodname{} rapidly learns successful policies from the shared dexterous prior, reaching high success within 2-5 hours. In contrast, training from scratch fails to make progress with either sparse task rewards or hand-engineered dense rewards.
  }
  \label{fig:compareWithScratch}
  \vspace{3mm}
\end{figure}

\begin{figure}[t!]
  \centering
  \includegraphics[width=\textwidth]{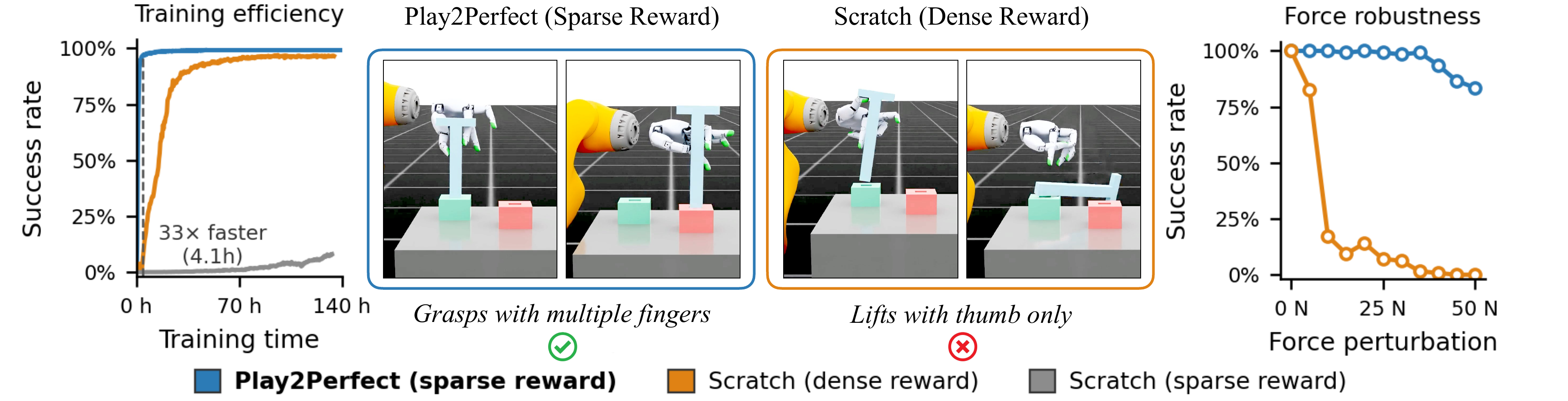}
  \caption{
    \textbf{Dexterous Play Pretraining Induces Robust Assembly Strategies.} After simplifying the initialization with an easy-grasp fixture, Scratch (dense reward) can learn the task, but it relies on a brittle strategy that balances the object rather than robustly grasping it. This shortcut fails sharply under external force perturbations. In contrast, \methodname{} learns a more stable grasping and recovery strategy, maintaining high success across perturbation magnitudes. 
  }
  \label{fig:compareWithScratchRobustness}
\end{figure}

\textbf{Setup.}
We first test whether a general dexterous prior improves RL learning compared to training from scratch. We measure success rate as a function of RL wall-clock time and compare \methodname{} against two scratch baselines: \textit{Scratch (sparse reward)}, which uses the same sparse assembly success reward as \methodname{}, and \textit{Scratch (dense reward)}, which additionally receives task-specific reward shaping for grasping, lifting, and tracking a sequence of 10 waypoints from the initial part pose to the fixture. In addition to the four main tasks, we include a simplified \texttt{Tight-Insertion (Fixtured)} task, where the T-peg starts propped up on a fixture. Computational resources and training budgets are detailed in Appendix~\ref{sec:simulation-compute}.

\textbf{Results.} Fig.~\ref{fig:compareWithScratch} shows that \methodname{} solves all tasks within roughly 2-5 hours of wall-clock RL training, whereas both scratch baselines produce no successful rollouts even after 24 hours. On the simplified \texttt{Tight-Insertion (Fixtured)} task, scratch training becomes feasible, as shown in Fig.~\ref{fig:compareWithScratchRobustness}. However, \textit{Scratch (dense reward)} requires over 100 hours to reach near-perfect success, while \methodname{} reaches the same success rate in only 4 hours, yielding a 33$\times$ speed-up. Moreover, the policy learned by \textit{Scratch (dense reward)} remains brittle: it balances the peg using the thumb rather than forming a stable grasp. Under external force perturbations, its success rate drops to $\sim$20\% with a 10N perturbation and eventually to 0\% under larger perturbations. In contrast, \methodname{} maintains over 75\% success even under the largest perturbations, indicating that \emph{play} pretraining induces a more robust manipulation strategy.

\Needspace{8\baselineskip}  
\subsection{Which Design Choices in \emph{Play} Pretraining Matter Most for Downstream Assembly?}

\textbf{Setup.}
We next ablate key \emph{play} pretraining choices and measure how they affect downstream RL finetuning on all 4 tasks averaged over 3 seeds. Per-task results and ablation implementation details are provided in Appendices~\ref{sec:additional-ablation-results} and~\ref{sec:pretraining-ablation-implementations}, respectively. We vary: 1) \emph{Object Diversity:} pretraining on 10, 100, or 1000 objects sampled from the same primitive distribution, with 1000 as \methodname{}'s default; 2) \emph{Training Objective:} comparing \methodname{}'s full 6D goal-pose objective against \textit{Translation-only}, which rewards only translational error along the same goal-pose sequences, and \textit{Rotation-only}, which first lifts the object to a random 6D pose and then samples only rotational goals at a fixed position; 3) \emph{Trajectory Diversity:} we compare using random goal trajectories (ours) against fixed sets of 10 or 100 trajectories; and 4) \emph{Goal Precision:} varying the success threshold $\epsilon$ for goal-reaching across 1cm (ours), 5cm, and 10cm.

\begin{figure}[t!]
  \centering
  \includegraphics[width=\textwidth]{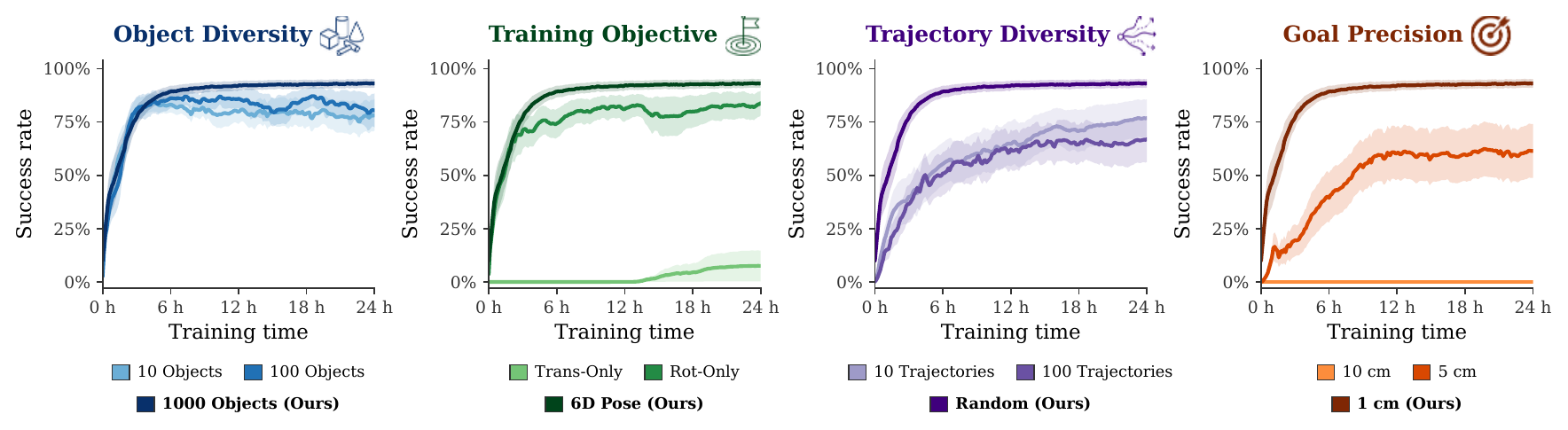}
\vspace{-5mm}

  \caption{
  \textbf{What Matters in Pretraining for Downstream Assembly Finetuning?}
We vary key \emph{play} pretraining choices and evaluate downstream RL finetuning success averaged across four assembly tasks and three seeds. Pretraining transfers best when it encourages \textbf{in-hand manipulation} via 6D in-hand object control across diverse objects and trajectories with precise goal tolerances.
  }
\vspace{-3mm}
  \label{fig:pretrainingAblations}
\end{figure}

\textbf{Results.}
Fig.~\ref{fig:pretrainingAblations} shows that all four design choices affect downstream assembly finetuning. 1) \emph{Object Diversity:} Training with more objects during pretraining leads to more stable RL finetuning, with final success rates consistently increasing from 10 to 1000 objects. 2) \emph{Training Objective:} Orientation control is critical. \textit{Translation-only} pretraining learns grasping and lifting, but does not learn object orientation control, and therefore fails to provide the in-hand reorientation prior needed for assembly. \textit{Rotation-only} pretraining transfers well, but plateaus about 10 points below full 6D pose-goal pretraining, likely because it provides less practice coupling reorientation with translational object motion. 3) \emph{Trajectory Diversity:} Fixed sets of 10 and 100 trajectories perform similarly, while online random trajectories learn fastest  and plateau highest, suggesting that broader coverage of goal-pose transitions better matches downstream assembly finetuning. 4) \emph{Goal Precision:} Precise goals are important. A loose 10cm threshold fails to transfer because coarse goal reaching does not require accurate object-pose control. A 5\,cm threshold learns more slowly and plateaus near 60\%, indicating that precise \emph{play} induces priors better matched to tight-clearance assembly.

\subsection{Is RL Finetuning Necessary for Precise, Contact-Rich Assembly?}

\begin{figure}[t!]
  \centering
  \includegraphics[width=\textwidth]{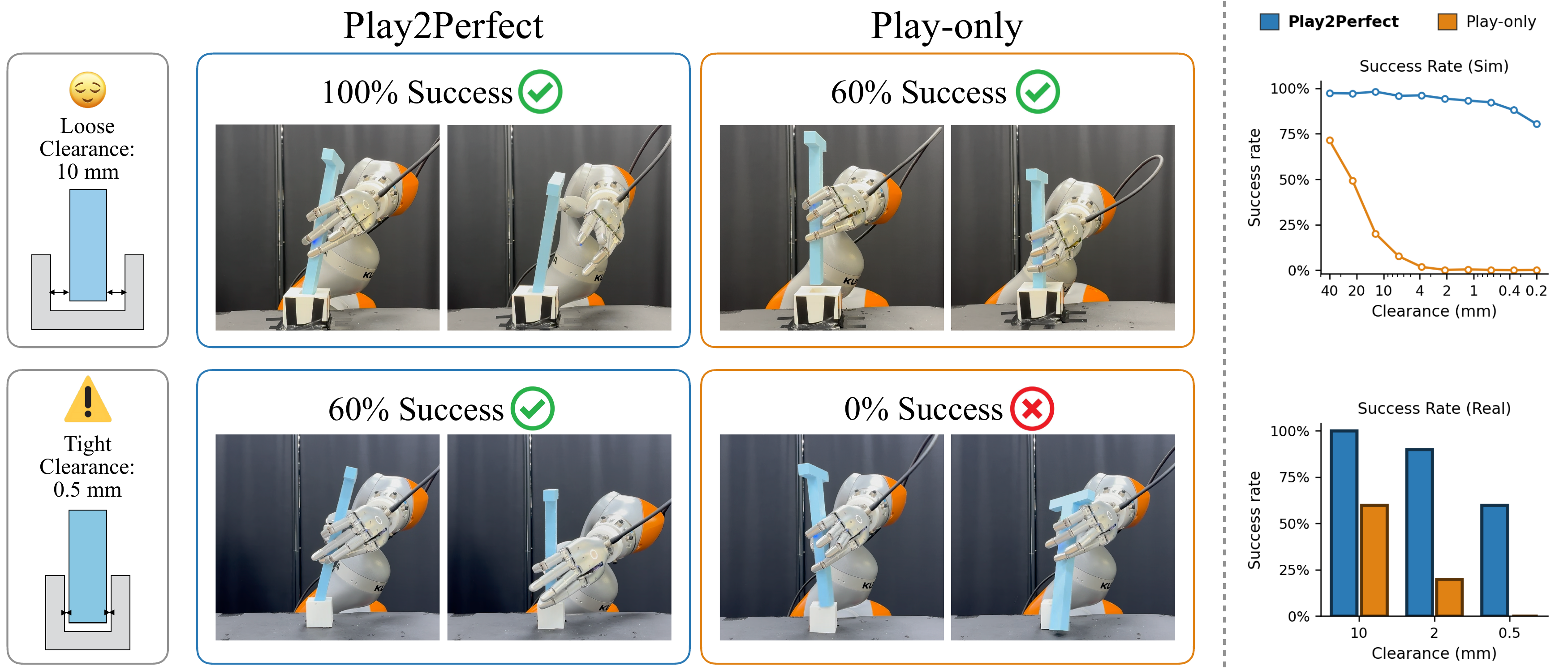}
  \caption{ 
    \textbf{Assembly Finetuning Enables Tight Insertion.} We compare \methodname{} against a frozen \textit{Play-only} policy on insertion tasks with varying contact clearance. (Left) Both policies succeed at loose clearance, but only \methodname{} succeeds at tight clearance. (Top right) Simulation sweeps show that \methodname{} remains robust as clearance decreases, while \textit{Play-only} rapidly degrades. (Bottom right) Real-world success rates show the same trend across different clearances.
  }
  \label{fig:pegInHole}
  \vspace{-2mm}  
\end{figure}

\textbf{Setup.}
We next evaluate whether \emph{play} pretraining alone is sufficient for precise assembly, or whether task-specific RL finetuning is necessary. We compare \methodname{} against \textit{Play-only}, the pretrained policy without assembly finetuning. We evaluate both policies on \texttt{Tight-Insertion} across fixture holes with increasingly tight contact clearances. Clearance is defined as the difference between the hole and peg cross-sectional dimensions: for example, our peg has a $30\mathrm{mm}\times20\mathrm{mm}$ cross-section, so a 1mm clearance corresponds to a $31\mathrm{mm}\times21\mathrm{mm}$ hole. We finetune \methodname{} on 5--0.5 mm clearances, then evaluate both policies from loose 40 mm to tight 0.2 mm clearance.

\textbf{Results.}
Fig.~\ref{fig:pegInHole} shows that \textit{Play-only} solves only the loosest insertion settings. In simulation, it reaches 75\% success at 40 mm clearance but drops to nearly 0\% by 4 mm. In contrast, \methodname{} maintains high success as precision increases, achieving 95\% at 4 mm, 92\% at 1 mm, and 80\% at 0.2 mm, which is tighter than the training distribution. This shows that \emph{play} pretraining provides useful grasping and reorientation behaviors, but RL finetuning is still needed to turn this prior into a precise, contact-rich assembly policy. The real-world results show the same trend. At 10 mm clearance, \methodname{} achieves 100\% success compared to 60\% for \textit{Play-only}. At 2 mm clearance, \methodname{} achieves 90\% success while \textit{Play-only} drops to 20\%. At 0.5 mm clearance, \methodname{} still succeeds 60\% of the time, whereas \textit{Play-only} fails completely. Qualitatively, \textit{Play-only} tends to move directly toward the goal pose and treats contact as a disturbance. In contrast, \methodname{} learns to search locally near the hole, make corrective motions under contact, and commit to insertion once the part is aligned. 

\subsection{Can \methodname{} Transfer to the Real World?}
\begin{table}[t]
\centering
\footnotesize
\setlength{\tabcolsep}{5pt}  
\renewcommand{\arraystretch}{1.2}

\caption{
\textbf{Real-World Assembly Results.}
\methodname{} transfers zero-shot to real-world insertion, multi-part assembly, and screwing. Completion times are mean $\pm$ std over successful trials, including grasping, transport, and assembly (* indicates success conditional on the preceding sub-task).
}
\label{tab:real-world-results}
\vspace{0.5em}  

\begin{tabular}{lccccccc}
\toprule
& \multicolumn{3}{c}{\texttt{Tight-Insertion}}
& \multicolumn{2}{c}{\texttt{Assemble-Beam}}
& \multicolumn{2}{c}{\texttt{Screw-Leg}} \\
\cmidrule(lr){2-4}\cmidrule(lr){5-6}\cmidrule(lr){7-8}
& 10mm & 2mm & 0.5mm & Step 1 & Step 2* & Insert & Screw* \\
\midrule
Success Rate
& 10/10 & 9/10 & 6/10 & 16/20 & 12/16 & 7/10 & 5/7 \\
Completion Time
& 6.8 $\pm$ 1.5 s& 9.4 $\pm$ 1.9 s& 11.1 $\pm$ 5.1 s& \multicolumn{2}{c}{21.0 $\pm$ 8.9 s}& \multicolumn{2}{c}{15.6 $\pm$ 2.9 s} \\
\bottomrule
\end{tabular}
\vspace{-1.3em}  

\end{table}

\textbf{Setup.} We evaluate zero-shot sim-to-real transfer of \methodname{} on \texttt{Tight-Insertion}, \texttt{Assemble-Beam}, and \texttt{Screw-Leg}, using FoundationPose~\cite{wen2024foundationpose} for object pose tracking without real-world finetuning. For multi-step \texttt{Assemble-Beam}, we sequentially execute the policies for each assembly step without resetting the intermediate state. More details in Appendix~\ref{app:endtoendassembly}.

\textbf{Results.}
Table~\ref{tab:real-world-results} shows that \methodname{} transfers to the real world across all three assembly tasks. On \texttt{Tight-Insertion}, the policy maintains high success as clearance tightens, achieving 10/10 at 10 mm, 9/10 at 2 mm, and 6/10 at 0.5 mm. Completion time increases from $6.8 \pm 1.5$ s to $11.1 \pm 5.1$ s as the policy performs additional local search for tighter alignment. On \texttt{Assemble-Beam}, both steps succeed with high success, with 16/20 success on Step 1 and 12/16 on Step 2. On \texttt{Screw-Leg}, \methodname{} achieves 7/10 success on insertion and 5/7 on the subsequent screwing, with successful trials taking $15.6 \pm 2.9$ s including both phases. Completion time measures the full task duration, including approaching the object from the home position, grasping it, reorienting and transporting it, and finally executing the contact-rich assembly interaction. These fast execution times demonstrate the advantage of using dexterous hands for assembly, as well as the ability of RL to discover efficient manipulation strategies. Most failures occur during final contact-rich interactions, where occlusions can degrade perception and contact dynamics can introduce sim-to-real mismatch. We further investigate robustness to pose-estimation noise and characterize real-world failures (Appendices~\ref{app:pose_noise_failures}, \ref{app:pose_failures}, and \ref{app:experiment_analysis}). In simulation, the policies tolerate pose-observation noise beyond the training range, although \texttt{Screw-Leg} is the most sensitive. Across 60 real-world rollouts, we identify 18 failures, including timeouts (6), object drops during insertion (5) or reorientation (3), and catastrophic pose-tracking failures (4). Smaller pose-estimation errors may also contribute to other failures but cannot be reliably isolated without ground-truth poses.

\subsection{Does \methodname{} Transfer to Everyday Insertion Tasks?}
\begin{figure}[h]
  \centering
  \vspace{-2mm}
  \includegraphics[width=\linewidth]{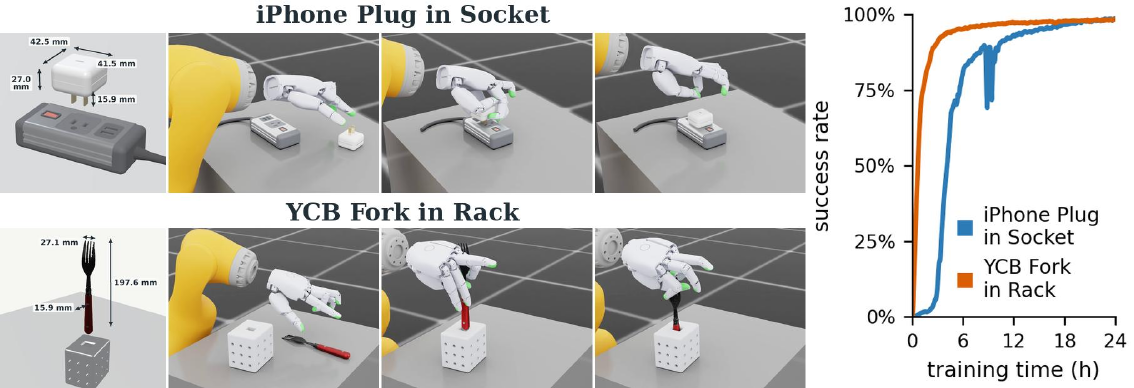}
\caption{\textbf{Transfer to Everyday Insertion Tasks.} \methodname{} achieves 90\% success in simulation on iPhone plug and YCB fork insertion after just 8.4 and 2.8 hours of finetuning, respectively. We use the same pretrained checkpoint and training recipe as the main experiments.}
  \label{fig:everyday-objects}
  \vspace{-1mm}
\end{figure}

\textbf{Setup.} We demonstrate the application of \methodname{} beyond structured assembly benchmarks to everyday insertion tasks involving smaller, more challenging object geometries. We consider two additional tasks: inserting an iPhone plug into a socket and a YCB fork into a narrow rack opening. Both tasks use the same pretrained play checkpoint and RL finetuning recipe as our main experiments, without task-specific modifications.

\textbf{Results.} Fig.~\ref{fig:everyday-objects} shows that \methodname{} successfully adapts to both tasks, achieving 90\% success in simulation after 8.4 and 2.8 hours of finetuning for the iPhone plug and YCB fork, respectively. These results demonstrate that the learned dexterous prior extends beyond structured industrial assembly to everyday insertion tasks with diverse object geometries.

\section{Discussion and Limitations}
\label{sec:conclusion}
We presented \methodname{}, a framework for precise contact-rich assembly that finetunes a dexterous prior learned through play. By \emph{play} pretraining across diverse objects, \methodname{} enables fast RL on assembly tasks by adapting the prior to precise contact interactions. Our ablations show that object diversity, objectives, trajectory diversity, and goal precision all affect downstream performance. Finally, \methodname{} transfers zero-shot to challenging real-world dexterous assembly tasks.

\textbf{Limitations.} Our system learns short-horizon assembly skills rather than a complete autonomous assembly pipeline. Task sequencing, active-part selection, and goal poses are specified externally, and policies are finetuned per task or benchmark family. Future work could combine these skills with sequencing, scene memory, recovery, and broader multi-task finetuning. Real-world deployment also depends on object-pose estimates, which can fail under occlusion or fast motion. Finally, beyond the goal pose, the policy does not directly observe the fixture or surrounding geometry. Incorporating visual or tactile observations could address these perception and scene-awareness limitations.

\section*{Acknowledgements}

This work is supported by Stanford Human-Centered Artificial
Intelligence (HAI), ONR Young Investigator Award, the National
Science Foundation (NSF) under Grant Numbers 2153854, 2327974,
2312956, 2327973, and 2342246, and the Natural Sciences and
Engineering Research Council of Canada (NSERC) under Award Number
526541680. We thank Sharpa for their research collaboration and for the technical support provided by their team, specifically Kaifeng Zhang, Wenjie Mei, Yi Zhou, Yunfang Yang, Jie Yin, Jason Lee, and Wanli Xing.



\bibliography{refs, simtoolreal_refs}  

@article{tian2022assemble,
  title={Assemble them all: Physics-based planning for generalizable assembly by disassembly},
  author={Tian, Yunsheng and Xu, Jie and Li, Yichen and Luo, Jieliang and Sueda, Shinjiro and Li, Hui and Willis, Karl DD and Matusik, Wojciech},
  journal={ACM Transactions on Graphics (TOG)},
  volume={41},
  number={6},
  pages={1--11},
  year={2022},
  publisher={ACM New York, NY, USA}
}

@inproceedings{lynch2020learning,
  title={Learning latent plans from play},
  author={Lynch, Corey and Khansari, Mohi and Xiao, Ted and Kumar, Vikash and Tompson, Jonathan and Levine, Sergey and Sermanet, Pierre},
  booktitle={Conference on robot learning},
  pages={1113--1132},
  year={2020},
  organization={Pmlr}
}

@article{wang2023mimicplay,
  title={Mimicplay: Long-horizon imitation learning by watching human play},
  author={Wang, Chen and Fan, Linxi and Sun, Jiankai and Zhang, Ruohan and Fei-Fei, Li and Xu, Danfei and Zhu, Yuke and Anandkumar, Anima},
  journal={arXiv preprint arXiv:2302.12422},
  year={2023}
}

@article{tao2025dexwild,
  title={Dexwild: Dexterous human interactions for in-the-wild robot policies},
  author={Tao, Tony and Srirama, Mohan Kumar and Liu, Jason Jingzhou and Shaw, Kenneth and Pathak, Deepak},
  journal={arXiv preprint arXiv:2505.07813},
  year={2025}
}

@article{qiu2025humanoid,
  title={Humanoid policy\~{} human policy},
  author={Qiu, Ri-Zhao and Yang, Shiqi and Cheng, Xuxin and Chawla, Chaitanya and Li, Jialong and He, Tairan and Yan, Ge and Yoon, David J and Hoque, Ryan and Paulsen, Lars and others},
  journal={arXiv preprint arXiv:2503.13441},
  year={2025}
}

@article{yang2025egovla,
  title={Egovla: Learning vision-language-action models from egocentric human videos},
  author={Yang, Ruihan and Yu, Qinxi and Wu, Yecheng and Yan, Rui and Li, Borui and Cheng, An-Chieh and Zou, Xueyan and Fang, Yunhao and Cheng, Xuxin and Qiu, Ri-Zhao and others},
  journal={arXiv preprint arXiv:2507.12440},
  year={2025}
}

@article{zheng2026egoscale,
  title={Egoscale: Scaling dexterous manipulation with diverse egocentric human data},
  author={Zheng, Ruijie and Niu, Dantong and Xie, Yuqi and Wang, Jing and Xu, Mengda and Jiang, Yunfan and Casta{\~n}eda, Fernando and Hu, Fengyuan and Tan, You Liang and Fu, Letian and others},
  journal={arXiv preprint arXiv:2602.16710},
  year={2026}
}

@article{khazatsky2024droid,
  title={Droid: A large-scale in-the-wild robot manipulation dataset},
  author={Khazatsky, Alexander and Pertsch, Karl and Nair, Suraj and Balakrishna, Ashwin and Dasari, Sudeep and Karamcheti, Siddharth and Nasiriany, Soroush and Srirama, Mohan Kumar and Chen, Lawrence Yunliang and Ellis, Kirsty and others},
  journal={arXiv preprint arXiv:2403.12945},
  year={2024}
}

@inproceedings{o2024open,
  title={Open x-embodiment: Robotic learning datasets and rt-x models: Open x-embodiment collaboration 0},
  author={O’Neill, Abby and Rehman, Abdul and Maddukuri, Abhiram and Gupta, Abhishek and Padalkar, Abhishek and Lee, Abraham and Pooley, Acorn and Gupta, Agrim and Mandlekar, Ajay and Jain, Ajinkya and others},
  booktitle={2024 IEEE International Conference on Robotics and Automation (ICRA)},
  pages={6892--6903},
  year={2024},
  organization={IEEE}
}

@article{barreiros2026careful,
  title={A careful examination of large behavior models for multitask dexterous manipulation},
  author={Barreiros, Jose and Beaulieu, Andrew and Bhat, Aditya and Cory, Rick and Cousineau, Eric and Dai, Hongkai and Fang, Ching-Hsin and Hashimoto, Kunimatsu and Irshad, Muhammad Zubair and Itkina, Masha and others},
  journal={Science Robotics},
  volume={11},
  number={113},
  pages={eaea6201},
  year={2026},
  publisher={American Association for the Advancement of Science}
}

@article{nvidia2025gr00t,
  title={Gr00t n1: An open foundation model for generalist humanoid robots},
  author={Nvidia, J Bjorck and Castaneda, Fernando and Cherniadev, N and Da, X and Ding, R and Fan, L and Fang, Y and Fox, D and Hu, F and Huang, S and others},
  journal={arXiv preprint arXiv:2503.14734},
  volume={2},
  year={2025}
}

@article{intelligence2025pi05,
  title={{$\pi_{0.5}$}: A Vision-Language-Action Model with Open-World Generalization},
  author={{Physical Intelligence} and Black, Kevin and Brown, Noah and Darpinian, James and Dhabalia, Karan and Driess, Danny and Esmail, Adnan and Equi, Michael and Finn, Chelsea and Fusai, Niccolo and others},
  journal={arXiv preprint arXiv:2504.16054},
  year={2025}
}

@article{si2025exostart,
  title={ExoStart: Efficient learning for dexterous manipulation with sensorized exoskeleton demonstrations},
  author={Si, Zilin and Chen, Jose Enrique and Karagozler, M Emre and Bronars, Antonia and Hutchinson, Jonathan and Lampe, Thomas and Gileadi, Nimrod and Howell, Taylor and Saliceti, Stefano and Barczyk, Lukasz and others},
  journal={arXiv preprint arXiv:2506.11775},
  year={2025}
}

@inproceedings{bauza2025demostart,
  title={Demostart: Demonstration-led auto-curriculum applied to sim-to-real with multi-fingered robots},
  author={Bauza, Maria and Chen, Jose Enriaue and Dalibard, Valentin and Gileadi, Nimrod and Hafner, Roland and Martins, Murilo F and Moore, Joss and Pevceviciute, Rugile and Laurens, Antoine and Rao, Dushyant and others},
  booktitle={2025 IEEE International Conference on Robotics and Automation (ICRA)},
  pages={6756--6763},
  year={2025},
  organization={IEEE}
}

@article{chen2023sequential,
  title={Sequential dexterity: Chaining dexterous policies for long-horizon manipulation},
  author={Chen, Yuanpei and Wang, Chen and Fei-Fei, Li and Liu, C Karen},
  journal={arXiv preprint arXiv:2309.00987},
  year={2023}
}

@article{wang2024dexcap,
  title={Dexcap: Scalable and portable mocap data collection system for dexterous manipulation},
  author={Wang, Chen and Shi, Haochen and Wang, Weizhuo and Zhang, Ruohan and Fei-Fei, Li and Liu, C Karen},
  journal={arXiv preprint arXiv:2403.07788},
  year={2024}
}

@article{cheng2024open,
  title={Open-television: Teleoperation with immersive active visual feedback},
  author={Cheng, Xuxin and Li, Jialong and Yang, Shiqi and Yang, Ge and Wang, Xiaolong},
  journal={arXiv preprint arXiv:2407.01512},
  year={2024}
}

@article{qin2023anyteleop,
  title={Anyteleop: A general vision-based dexterous robot arm-hand teleoperation system},
  author={Qin, Yuzhe and Yang, Wei and Huang, Binghao and Van Wyk, Karl and Su, Hao and Wang, Xiaolong and Chao, Yu-Wei and Fox, Dieter},
  journal={arXiv preprint arXiv:2307.04577},
  year={2023}
}

@inproceedings{arunachalam2023dexterous,
  title={Dexterous imitation made easy: A learning-based framework for efficient dexterous manipulation},
  author={Arunachalam, Sridhar Pandian and Silwal, Sneha and Evans, Ben and Pinto, Lerrel},
  booktitle={2023 ieee international conference on robotics and automation (icra)},
  pages={5954--5961},
  year={2023},
  organization={IEEE}
}

@inproceedings{lin2025learning,
  title={Learning visuotactile skills with two multifingered hands},
  author={Lin, Toru and Zhang, Yu and Li, Qiyang and Qi, Haozhi and Yi, Brent and Levine, Sergey and Malik, Jitendra},
  booktitle={2025 IEEE International Conference on Robotics and Automation (ICRA)},
  pages={5637--5643},
  year={2025},
  organization={IEEE}
}

@InProceedings{wen2024foundationpose,
  author    = {Wen, Bowen and Yang, Wei and Kautz, Jan and Birchfield, Stan},
  title     = {FoundationPose: Unified 6D Pose Estimation and
  Tracking of Novel Objects},
  booktitle = {Proceedings of the IEEE/CVF Conference on Computer
  Vision and Pattern Recognition (CVPR)},
  month     = {June},
  year      = {2024},
  pages     = {17868-17879}
}

@article{kuang2026dex4d,
  title={Dex4D: Task-Agnostic Point Track Policy for Sim-to-Real Dexterous Manipulation},
  author={Kuang, Yuxuan and Park, Sungjae and Fragkiadaki, Katerina and Tulsiani, Shubham},
  journal={arXiv preprint arXiv:2602.15828},
  year={2026}
}

@article{andrychowicz2020learning,
  title={Learning dexterous in-hand manipulation},
  author={Andrychowicz, OpenAI: Marcin and Baker, Bowen and Chociej, Maciek and Jozefowicz, Rafal and McGrew, Bob and Pachocki, Jakub and Petron, Arthur and Plappert, Matthias and Powell, Glenn and Ray, Alex and others},
  journal={The International Journal of Robotics Research},
  volume={39},
  number={1},
  pages={3--20},
  year={2020},
  publisher={SAGE Publications Sage UK: London, England}
}

@article{chen2023visual,
  title={Visual dexterity: In-hand reorientation of novel and complex object shapes},
  author={Chen, Tao and Tippur, Megha and Wu, Siyang and Kumar, Vikash and Adelson, Edward and Agrawal, Pulkit},
  journal={Science Robotics},
  volume={8},
  number={84},
  pages={eadc9244},
  year={2023},
  publisher={American Association for the Advancement of Science}
}

@inproceedings{zhang2024dexgraspnet,
  title={Dexgraspnet 2.0: Learning generative dexterous grasping in large-scale synthetic cluttered scenes},
  author={Zhang, Jialiang and Liu, Haoran and Li, Danshi and Yu, XinQiang and Geng, Haoran and Ding, Yufei and Chen, Jiayi and Wang, He},
  booktitle={8th Annual Conference on Robot Learning},
  year={2024}
}

@inproceedings{wan2023unidexgrasp++,
  title={Unidexgrasp++: Improving dexterous grasping policy learning via geometry-aware curriculum and iterative generalist-specialist learning},
  author={Wan, Weikang and Geng, Haoran and Liu, Yun and Shan, Zikang and Yang, Yaodong and Yi, Li and Wang, He},
  booktitle={Proceedings of the IEEE/CVF International Conference on Computer Vision},
  pages={3891--3902},
  year={2023}
}

@inproceedings{handa2023dextreme,
  title={Dextreme: Transfer of agile in-hand manipulation from simulation to reality},
  author={Handa, Ankur and Allshire, Arthur and Makoviychuk, Viktor and Petrenko, Aleksei and Singh, Ritvik and Liu, Jingzhou and Makoviichuk, Denys and Van Wyk, Karl and Zhurkevich, Alexander and Sundaralingam, Balakumar and others},
  booktitle={2023 IEEE International Conference on Robotics and Automation (ICRA)},
  pages={5977--5984},
  year={2023},
  organization={IEEE}
}

@article{lin2024twisting,
  title={Twisting lids off with two hands},
  author={Lin, Toru and Yin, Zhao-Heng and Qi, Haozhi and Abbeel, Pieter and Malik, Jitendra},
  journal={arXiv preprint arXiv:2403.02338},
  year={2024}
}

@inproceedings{shao2020learning,
  title={Learning to scaffold the development of robotic manipulation skills},
  author={Shao, Lin and Migimatsu, Toki and Bohg, Jeannette},
  booktitle={2020 IEEE International Conference on Robotics and Automation (ICRA)},
  pages={5671--5677},
  year={2020},
  organization={IEEE}
}

@inproceedings{tian2025fabrica,
  title={Fabrica: Dual-Arm Assembly of General Multi-Part Objects via Integrated Planning and Learning},
  author={Yunsheng Tian and Joshua Jacob and Yijiang Huang and Jialiang Zhao and Edward Li Gu and Pingchuan Ma and Annan Zhang and Farhad Javid and Branden Romero and Sachin Chitta and Shinjiro Sueda and Hui Li and Wojciech Matusik},
  booktitle={9th Annual Conference on Robot Learning},
  year={2025},
  url={https://openreview.net/forum?id=aSUNzvEJIf}
}

@article{luo2024fmb,
  title={FMB: a Functional Manipulation Benchmark for Generalizable Robotic Learning},
  author={Luo, Jianlan and Xu, Charles and Liu, Fangchen and Tan, Liam and Lin, Zipeng and Wu, Jeffrey and Abbeel, Pieter and Levine, Sergey},
  journal={arXiv preprint arXiv:2401.08553},
  year={2024}
}

@inproceedings{heo2023furniturebench,
  title={FurnitureBench: Reproducible Real-World Benchmark for Long-Horizon Complex Manipulation},
  author={Minho Heo and Youngwoon Lee and Doohyun Lee and Joseph J. Lim},
  booktitle={Robotics: Science and Systems},
  year={2023}
}

@inproceedings{jiang2024transic,
  title     = {TRANSIC: Sim-to-Real Policy Transfer by Learning from Online Correction},
  author    = {Yunfan Jiang and Chen Wang and Ruohan Zhang and Jiajun Wu and Li Fei-Fei},
  booktitle = {Conference on Robot Learning},
  year      = {2024}
}

@inproceedings{ankile2025imitation,
  title={From imitation to refinement-residual rl for precise assembly},
  author={Ankile, Lars and Simeonov, Anthony and Shenfeld, Idan and Torne, Marcel and Agrawal, Pulkit},
  booktitle={2025 IEEE International Conference on Robotics and Automation (ICRA)},
  pages={01--08},
  year={2025},
  organization={IEEE}
}

@inproceedings{yin2026emergent,
  title={Emergent Dexterity via Diverse Resets and Large-Scale Reinforcement Learning},
  author={Patrick Yin and Tyler Westenbroek and Zhengyu Zhang and Joshua Tran and Ignacio Dagnino and Eeshani Shilamkar and Numfor Mbiziwo-Tiapo and Simran Bagaria and Xinlei Liu and Galen Mullins and Andrey Kolobov and Abhishek Gupta},
  booktitle={The Fourteenth International Conference on Learning Representations},
  year={2026},
  url={https://arxiv.org/abs/2603.15789},
}

@inproceedings{tang2024automate,
  title={AutoMate: Specialist and Generalist Assembly Policies over Diverse Geometries},
  author={Tang, Bingjie and Akinola, Iretiayo and Xu, Jie and Wen, Bowen and Handa, Ankur and Van Wyk, Karl and Fox, Dieter and Sukhatme, Gaurav S. and Ramos, Fabio and Narang, Yashraj},
  booktitle={Robotics: Science and Systems},
  year={2024}
}

@inproceedings{tang2023industreal,
  title={IndustReal: Transferring Contact-Rich Assembly Tasks from Simulation to Reality},
  author={Tang, Bingjie and Lin, Michael A. and Akinola, Iretiayo and Handa, Ankur and Sukhatme, Gaurav S. and Ramos, Fabio and Fox, Dieter and Narang, Yashraj S.},
  booktitle={Robotics: Science and Systems},
  year={2023}
}

@article{ankile2024juicer,
  author    = {Ankile, Lars and Simeonov, Anthony and Shenfeld, Idan and Agrawal, Pulkit},
  title     = {JUICER: Data-Efficient Imitation Learning for Robotic Assembly},
  journal   = {arXiv},
  year      = {2024},
}

@article{shi2023robocook,
  title={RoboCook: Long-Horizon Elasto-Plastic Object Manipulation with Diverse Tools},
  author={Shi, Haochen and Xu, Huazhe and Clarke, Samuel and Li, Yunzhu and Wu, Jiajun},
  journal={arXiv preprint arXiv:2306.14447},
  year={2023},
}

@inproceedings{ha2020fit2form,
  title={{Fit2Form}: 3{D} Generative Model for Robot Gripper Form Design},
  author={Ha, Huy and Agrawal, Shubham and Song, Shuran},
  booktitle={Conference on Robotic Learning (CoRL)},
  year={2020}
}

@misc{kedia2026simtoolrealobjectcentricpolicyzeroshot,
  title={SimToolReal: An Object-Centric Policy for Zero-Shot Dexterous Tool Manipulation},
  author={Kushal Kedia and Tyler Ga Wei Lum and Jeannette Bohg and C. Karen Liu},
  year={2026},
  eprint={2602.16863},
  archivePrefix={arXiv},
  primaryClass={cs.RO},
  url={https://arxiv.org/abs/2602.16863}
}

@misc{yin2025dexteritygen,
  title={DexterityGen: Foundation Controller for Unprecedented Dexterity},
  author={Zhao-Heng Yin and Changhao Wang and Luis Pineda and Francois Hogan and Krishna Bodduluri and Akash Sharma and Patrick Lancaster and Ishita Prasad and Mrinal Kalakrishnan and Jitendra Malik and Mike Lambeta and Tingfan Wu and Pieter Abbeel and Mustafa Mukadam},
  year={2025},
  eprint={2502.04307},
  archivePrefix={arXiv},
  primaryClass={cs.RO},
  url={https://arxiv.org/abs/2502.04307}
}

@inproceedings{sapg2024,
  title     = {SAPG: Split and Aggregate Policy Gradients},
  author    = {Singla, Jayesh and Agarwal, Ananye and Pathak, Deepak},
  booktitle = {Proceedings of the 41st International Conference on Machine Learning (ICML 2024)},
  year      = {2024},
  series    = {Proceedings of Machine Learning Research},
  address   = {Vienna, Austria},
  month     = {July},
  publisher = {PMLR}
}

@article{chen2021system,
  title={A System for General In-Hand Object Re-Orientation},
  author={Chen, Tao and Xu, Jie and Agrawal, Pulkit},
  journal={Conference on Robot Learning},
  year={2021}
}

@article{singh2024dextrah,
  title={Dextrah-rgb: Visuomotor policies to grasp anything with
  dexterous hands},
  author={Singh, Ritvik and Allshire, Arthur and Handa, Ankur and
  Ratliff, Nathan and Van Wyk, Karl},
  journal={arXiv preprint arXiv:2412.01791},
  year={2024}
}

@article{singh2025end,
  title={End-to-end RL Improves Dexterous Grasping Policies},
  author={Singh, Ritvik and Van Wyk, Karl and Abbeel, Pieter and
  Malik, Jitendra and Ratliff, Nathan and Handa, Ankur},
  journal={arXiv preprint arXiv:2509.16434},
  year={2025}
}

@inproceedings{li2025maniptrans,
  title={Maniptrans: Efficient dexterous bimanual manipulation
  transfer via residual learning},
  author={Li, Kailin and Li, Puhao and Liu, Tengyu and Li, Yuyang and
  Huang, Siyuan},
  booktitle={Proceedings of the Computer Vision and Pattern
  Recognition Conference},
  pages={6991--7003},
  year={2025}
}

@article{arunachalam2022holo,
  title={Holo-dex: Teaching dexterity with immersive mixed reality},
  author={Arunachalam, Sridhar Pandian and G{\"u}zey, Irmak and
  Chintala, Soumith and Pinto, Lerrel},
  journal={arXiv preprint arXiv:2210.06463},
  year={2022}
}

@article{sivakumar2022robotic,
  title={Robotic telekinesis: Learning a robotic hand imitator by
  watching humans on youtube},
  author={Sivakumar, Aravind and Shaw, Kenneth and Pathak, Deepak},
  journal={arXiv preprint arXiv:2202.10448},
  year={2022}
}

@article{iyer2024open,
  title={Open teach: A versatile teleoperation system for robotic manipulation},
  author={Iyer, Aadhithya and Peng, Zhuoran and Dai, Yinlong and
  Guzey, Irmak and Haldar, Siddhant and Chintala, Soumith and Pinto, Lerrel},
  journal={arXiv preprint arXiv:2403.07870},
  year={2024}
}

@article{pacchierotti2023cutaneous,
  title={Cutaneous/tactile haptic feedback in robotic teleoperation:
  Motivation, survey, and perspectives},
  author={Pacchierotti, Claudio and Prattichizzo, Domenico},
  journal={IEEE Transactions on Robotics},
  volume={40},
  pages={978--998},
  year={2023},
  publisher={IEEE}
}

@misc{handa2019dexpilotvisionbasedteleoperation,
  title={DexPilot: Vision Based Teleoperation of Dexterous Robotic
  Hand-Arm System},
  author={Ankur Handa and Karl Van Wyk and Wei Yang and Jacky Liang
    and Yu-Wei Chao and Qian Wan and Stan Birchfield and Nathan Ratliff
  and Dieter Fox},
  year={2019},
  eprint={1910.03135},
  archivePrefix={arXiv},
  primaryClass={cs.CV},
  url={https://arxiv.org/abs/1910.03135},
}

@misc{agarwal2023dexterousfunctionalgrasping,
  title={Dexterous Functional Grasping},
  author={Ananye Agarwal and Shagun Uppal and Kenneth Shaw and Deepak Pathak},
  year={2023},
  eprint={2312.02975},
  archivePrefix={arXiv},
  primaryClass={cs.RO},
  url={https://arxiv.org/abs/2312.02975},
}

@article{PPO,
  author = {Schulman, John and Wolski, Filip and Dhariwal, Prafulla
  and Radford, Alec and Klimov, Oleg},

  title = {Proximal Policy Optimization Algorithms},

  publisher = {arXiv},
  journal={arXiv preprint arXiv:1707.06347},
  year = {2017},

  copyright = {arXiv.org perpetual, non-exclusive license}
}

@inproceedings{lum2024dextrahg,

title     = {Dextr{AH}-G: Pixels-to-Action Dexterous Arm-Hand
Grasping with Geometric Fabrics},

author    = {Tyler Ga Wei Lum and Martin Matak and Viktor Makoviychuk
  and Ankur Handa and Arthur Allshire and Tucker Hermans and Nathan D.
Ratliff and Karl Van Wyk},

booktitle = {8th Annual Conference on Robot Learning},

year      = {2024},

url       = {https://openreview.net/forum?id=S2Jwb0i7HN}

}

@ARTICLE{chen2025dexforceextractingforceinformedactions,
author={Chen, Claire and Yu, Zhongchun and Choi, Hojung and Cutkosky,
Mark and Bohg, Jeannette},
journal={IEEE Robotics and Automation Letters},
title={DexForce: Extracting Force-Informed Actions From Kinesthetic
Demonstrations for Dexterous Manipulation},
year={2025},
volume={10},
number={6},
pages={6416-6423}
}

@misc{liu2025dexndmclosingrealitygap,
title={DexNDM: Closing the Reality Gap for Dexterous In-Hand Rotation
via Joint-Wise Neural Dynamics Model},
author={Xueyi Liu and He Wang and Li Yi},
year={2025},
eprint={2510.08556},
archivePrefix={arXiv},
primaryClass={cs.RO},
url={https://arxiv.org/abs/2510.08556},
}

@inproceedings{shaw2024bimanual,
title={Bimanual Dexterity for Complex Tasks},
author={Shaw, Kenneth and Li, Yulong and Yang, Jiahui and Srirama,
  Mohan Kumar and Liu, Ray and Xiong, Haoyu and Mendonca, Russell and
Pathak, Deepak},
booktitle={8th Annual Conference on Robot Learning},
year={2024}
}

@article{bunny-visionpro,
title   = {Bunny-VisionPro: Real-Time Bimanual Dexterous
Teleoperation for Imitation Learning},
author  = {Runyu Ding and Yuzhe Qin and Jiyue Zhu and Chengzhe Jia
and Shiqi Yang and Ruihan Yang and Xiaojuan Qi and Xiaolong Wang},
year    = {2024}
}

@article{fang2025dexop,
title={DEXOP: A Device for Robotic Transfer of Dexterous Human Manipulation},
author={Fang, Hao-Shu and Romero, Branden and Xie, Yichen and Hu,
  Arthur and Huang, Bo-Ruei and Alvarez, Juan and Kim, Matthew and
  Margolis, Gabriel and Anbarasu, Kavya and Tomizuka, Masayoshi and
Adelson, Edward and Agrawal, Pulkit},
journal={arXiv preprint arXiv:2509.04441},
year={2025}
}

@article{xu2025dexumi,
title={DexUMI: Using Human Hand as the Universal Manipulation
Interface for Dexterous Manipulation},
author={Xu, Mengda and Zhang, Han and Hou, Yifan and Xu, Zhenjia and
Fan, Linxi and Veloso, Manuela and Song, Shuran},
journal={arXiv preprint arXiv:2505.21864},
year={2025}
}

@inproceedings{ye2025dex1b,
title={Dex1B: Learning with 1B Demonstrations for Dexterous Manipulation},
author={Ye, Jianglong and Wang, Keyi and Yuan, Chengjing and Yang,
  Ruihan and Li, Yiquan and Zhu, Jiyue and Qin, Yuzhe and Zou, Xueyan
and Wang, Xiaolong},
booktitle={Robotics: Science and Systems (RSS)},
year={2025}
}

@misc{lum2025crossinghumanrobotembodimentgap,
author = {Tyler Ga Wei Lum and Olivia Y. Lee and C. Karen Liu and
Jeannette Bohg},
title = {Crossing the Human-Robot Embodiment Gap with Sim-to-Real RL
using One Human Demonstration},
year = {2025},
eprint = {2504.12609},
archivePrefix = {arXiv},
primaryClass = {cs.RO},
url = {https://arxiv.org/abs/2504.12609},
}

@misc{mandi2025dexmachinafunctionalretargetingbimanual,
title={DexMachina: Functional Retargeting for Bimanual Dexterous Manipulation},
author={Zhao Mandi and Yifan Hou and Dieter Fox and Yashraj Narang
and Ajay Mandlekar and Shuran Song},
year={2025},
eprint={2505.24853},
archivePrefix={arXiv},
primaryClass={cs.RO},
url={https://arxiv.org/abs/2505.24853},
}

@INPROCEEDINGS{Si-RSS-24,
AUTHOR    = {Zilin Si AND Kevin Lee Zhang AND Zeynep Temel AND Oliver Kroemer},
TITLE     = {{Tilde: Teleoperation for Dexterous In-Hand Manipulation
Learning with a DeltaHand}},
BOOKTITLE = {Proceedings of Robotics: Science and Systems},
YEAR      = {2024},
ADDRESS   = {Delft, Netherlands},
MONTH     = {July},
DOI       = {10.15607/RSS.2024.XX.128}
}

@INPROCEEDINGS{Human2RobotWholeBodyTransfer,
author={Arduengo, Miguel and Arduengo, Ana and Colomé, Adrià and
Lobo-Prat, Joan and Torras, Carme},
booktitle={2020 IEEE-RAS 20th International Conference on Humanoid
Robots (Humanoids)},
title={Human to Robot Whole-Body Motion Transfer},
year={2021},
volume={},
number={},
pages={299-305},
doi={10.1109/HUMANOIDS47582.2021.9555769}}

\clearpage

\appendix

\begin{center}
    {\Large \bfseries Appendix}
\end{center}

\section{Robustness to Noisy Pose Tracking and Failure Analysis}
\label{app:pose_noise_failures}
\begin{figure}[h]
  \begin{minipage}[t]{0.48\linewidth}
    \centering
    \includegraphics[width=\linewidth]{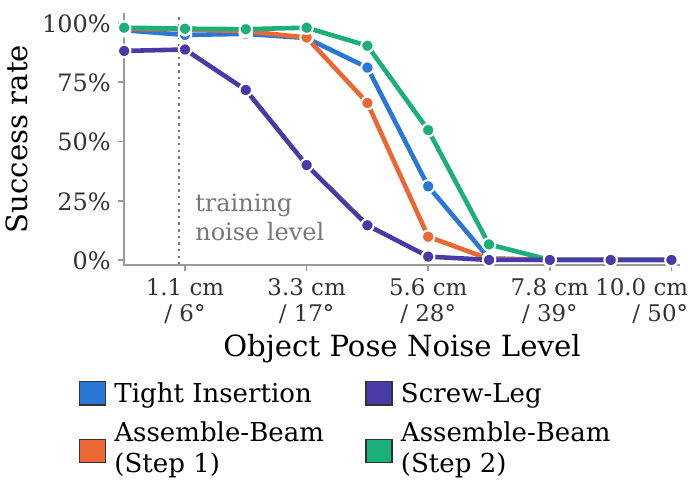}
    \caption{\textbf{Pose Noise Sensitivity (Simulation).} Success rates under increasing translational and rotational pose noise.}
    \label{fig:additional-pose-noise}
  \end{minipage}\hfill
  \begin{minipage}[t]{0.48\linewidth}
    \centering
    \includegraphics[width=\linewidth]{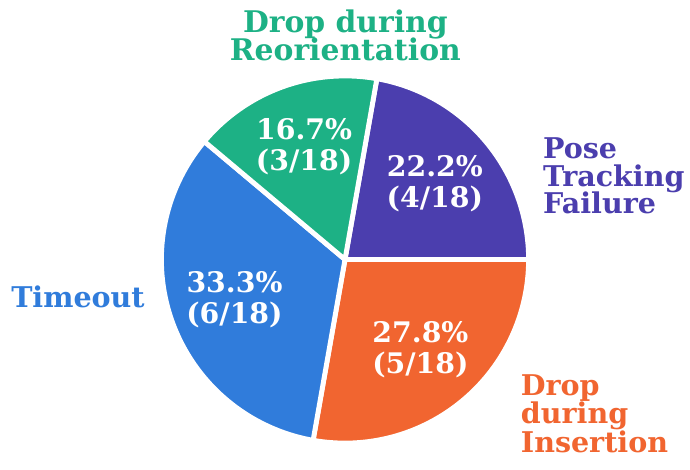}
    \caption{\textbf{Real-World Failure Analysis.} Failure modes and relative frequencies across real-world rollouts.}
    \label{fig:additional-failures}
  \end{minipage}
\end{figure}

We perform additional experiments to better understand our policies' robustness to pose noise. Fig.~\ref{fig:additional-pose-noise} evaluates policy success in simulation under increasing pose-observation noise across all four tasks. Although trained with 1cm translational and $5^\circ$ rotational noise, the policies tolerate noise beyond the training range, with \texttt{Screw-Leg} being most sensitive.

To better understand real-world failures, Fig.~\ref{fig:additional-failures} categorizes the 18 failures across 60 real-world trials spanning all four tasks. Catastrophic pose-tracking failures were easy to identify, as the pose tracker completely lost the object, but we could not reliably determine whether smaller tracking errors contributed to the remaining failures.

\section{End-to-End Two-Step Beam Assembly}
\label{app:endtoendassembly}
\begin{figure}[h]
  \centering
  \includegraphics[width=0.7\linewidth]{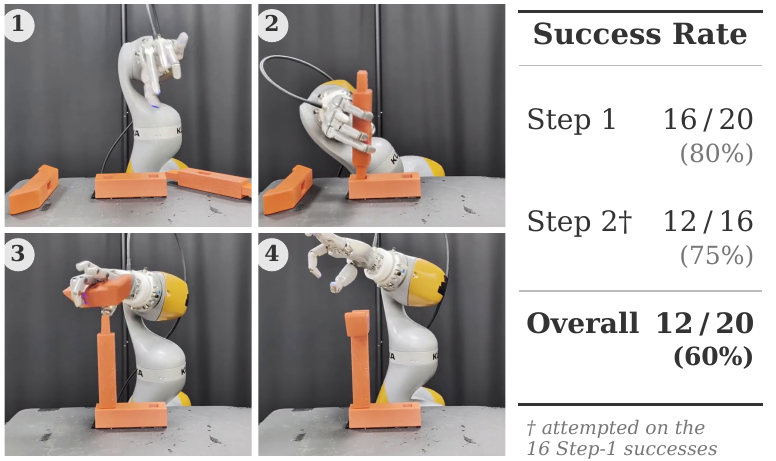}
  \caption{\textbf{End-to-End Beam Assembly.} Twenty real-world trials of both assembly steps in sequence, with four frames from a rollout. Step~2 success is conditional on Step~1 success.}
  \label{fig:additional-end-to-end}
\end{figure}

Here we provide more information about the end-to-end performance evaluation on the beam assembly task. Fig.~\ref{fig:additional-end-to-end} reports results over 20 real-world trials. If Step~1 succeeds, the Step~2 policy starts directly from the resulting assembly state. Step~1 succeeds in 16/20 trials, and Step~2 succeeds in 12/16 attempts following Step~1 success, giving an overall success rate of 12/20 (60\%).

\section{Additional Ablation Results}
\label{sec:additional-ablation-results}

The main paper reports ablation results averaged across all
tasks. Here, we provide the corresponding per-task results for
\texttt{Tight-Insertion}, \texttt{Assemble-Beam} Step 1,
\texttt{Assemble-Beam} Step 2, and \texttt{Screw-Leg}. For each task, we
repeat the same play pretraining ablations: object diversity, training
objective, trajectory diversity, and goal precision. Each pretrained policy is
then finetuned using the same sparse assembly reward and training procedure as
\methodname{}.

\begin{figure}[h] \centering \centerline{\small \texttt{Tight-Insertion}} \vspace{0.25em} \includegraphics[width=0.9\linewidth]{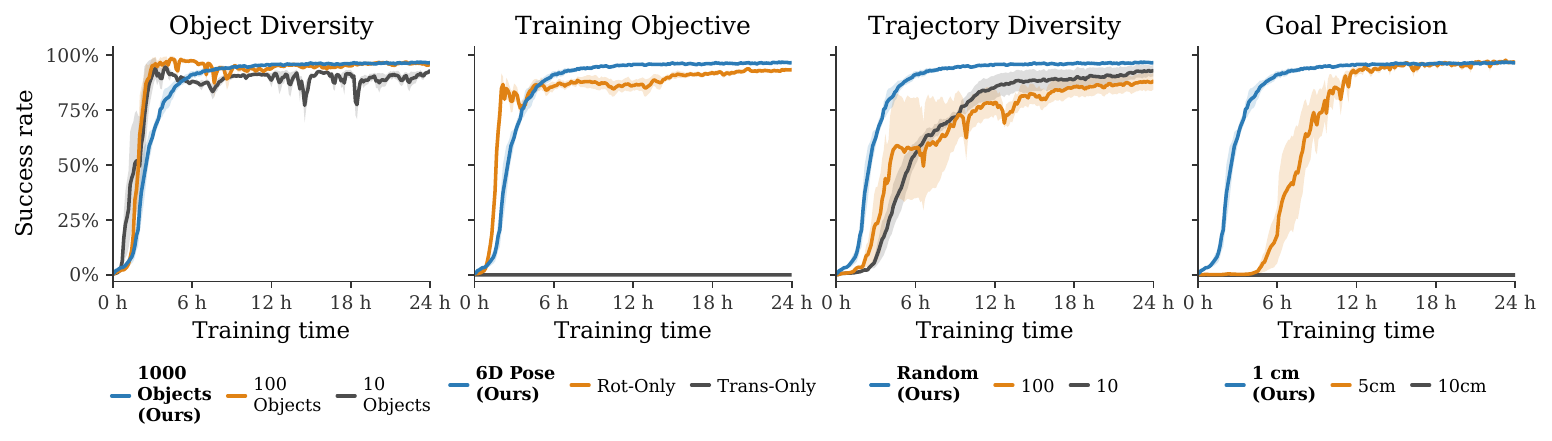} \vspace{0.75em} \centerline{\small \texttt{Assemble-Beam} Step 1} \vspace{0.25em} \includegraphics[width=0.9\linewidth]{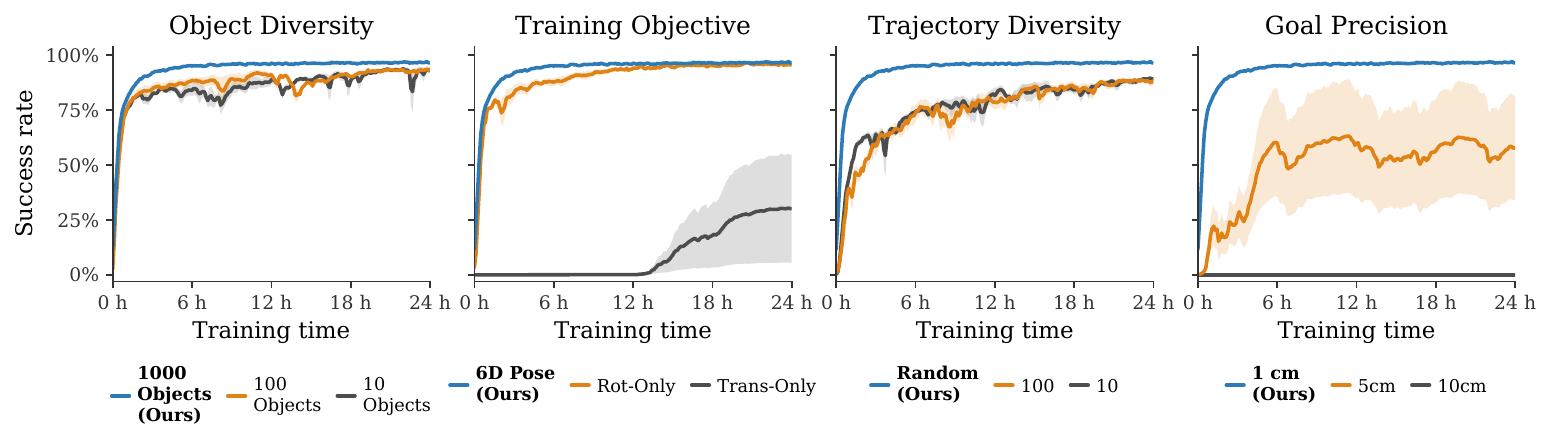} \vspace{0.75em} \centerline{\small \texttt{Assemble-Beam} Step 2} \vspace{0.25em} \includegraphics[width=0.9\linewidth]{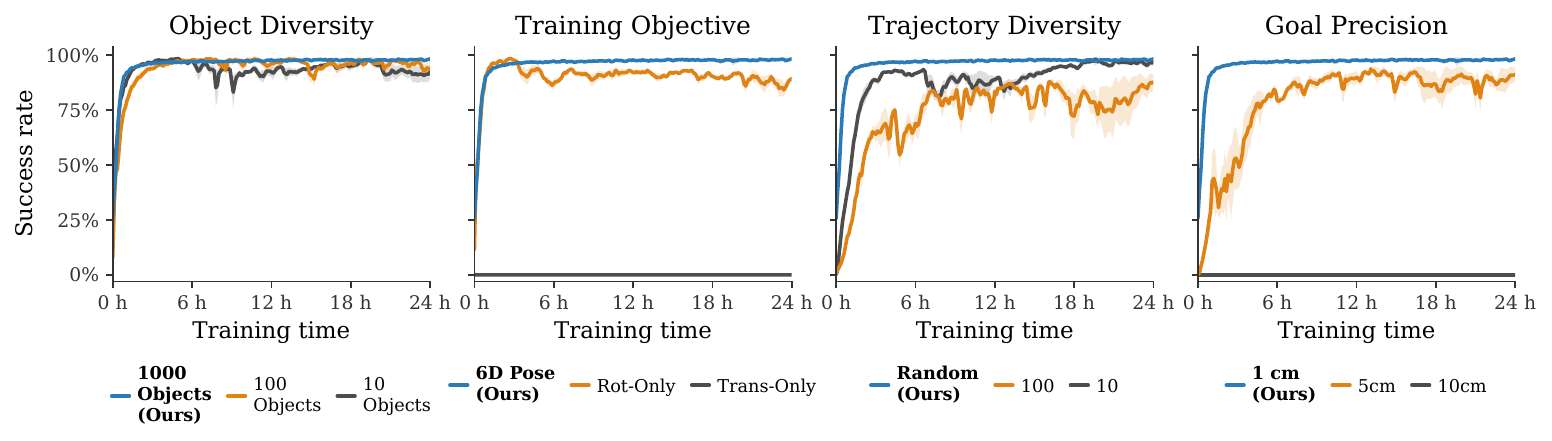} \vspace{0.75em} \centerline{\small \texttt{Screw-Leg}} \vspace{0.25em} \includegraphics[width=0.9\linewidth]{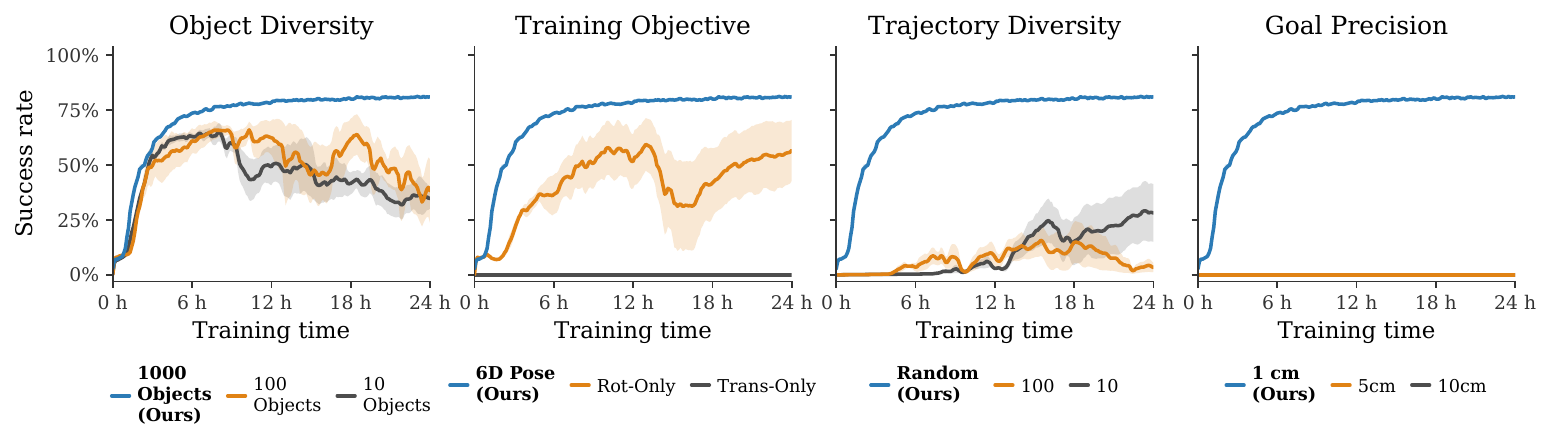} \caption{\textbf{Per-task pretraining ablation results.} We report the four play pretraining ablations from the main paper on each downstream assembly task. Across tasks, the results support the same conclusion as the averaged curves: pretraining transfers best when it teaches precise 6D in-hand object control across diverse objects and goal trajectories.} \label{fig:per-task-pretraining-ablations} \end{figure}

\textbf{Results.}
Fig.~\ref{fig:per-task-pretraining-ablations} shows that the trends observed in
the averaged ablation results are broadly consistent across individual
downstream tasks. Increasing \textbf{object diversity} generally improves
finetuning stability and final performance, indicating that the play prior
benefits from exposure to a broad distribution of object geometries and inertial
properties before being adapted to assembly. The \textbf{training objective} ablation shows that orientation control is
essential. Translation-only pretraining performs poorly across tasks because it
can be solved by grasping and transporting the object without learning the
in-hand reorientation skills needed for assembly. Rotation-only pretraining is
substantially stronger, but the full 6D pose objective is the most consistent
variant because it couples translation and rotation during play. The \textbf{trajectory diversity} ablation further shows that repeatedly
training on a small fixed set of goal trajectories can limit transfer. Online
random goal trajectories provide broader coverage of object-pose transitions,
which better matches the diversity of motions required during downstream
assembly finetuning.
Finally, the \textbf{goal precision} ablation shows that loose play objectives
do not reliably transfer to precise assembly. Larger goal tolerances can be
satisfied without accurate object-pose control, whereas the default 1 cm
tolerance forces the policy to acquire finer in-hand manipulation skills. Taken
together, the per-task results reinforce the central takeaway of the main
paper: effective play pretraining is not merely about learning to pick up and
move objects, but about learning precise finger-based 6D object control that can
be specialized to contact-rich assembly.

\section{Pose Tracking Failures with Smaller Objects}
\label{app:pose_failures}
\begin{figure}[h]
  \centering
  \includegraphics[width=0.65\linewidth]{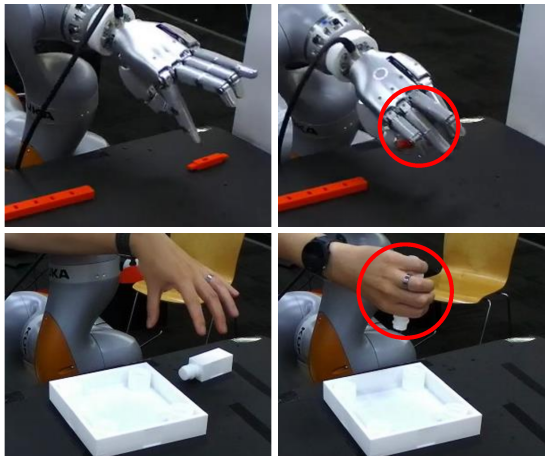}
  \caption{\textbf{Pose Tracking under Hand Occlusion.} At the original $1\times$ object scale, hand occlusion makes precise object pose tracking difficult. Enlarging the benchmark objects improves their visibility during real-world manipulation.}
  \label{fig:additional-occlusion}
\end{figure}

To illustrate the pose-tracking challenges that motivated enlarging the benchmark objects, Fig.~\ref{fig:additional-occlusion} shows examples of hand occlusion at the original object scale. We enlarged the beam parts by $3\times$ and made the screw leg $3\times$ longer primarily for reliable pose tracking in the real world, as the hand heavily occludes the smaller objects. The simulation results above evaluate smaller objects without this real-world tracking challenge.

\section{Real-World Experiment Additional Analysis}
\label{app:experiment_analysis}

\textbf{Qualitative behavior.}
Across real-world tasks, \methodname{} exhibits closed-loop recovery behaviors that are difficult to obtain from a single open-loop assembly motion. During insertion, the policy often searches locally near the hole, makes small corrective motions under contact, and commits once the part is better aligned. When the part becomes misaligned, the policy can recover with additional in-hand reorientations. If the part is dropped, the policy immediately regrasps and continues, provided the object remains within the workspace. In screwing tasks, the fingers spin the leg directly within the hand, enabling axial rotation while maintaining a stable grasp. With a parallel-jaw gripper, comparable reorientation would typically require slower extrinsic manipulation, such as placing and regrasping the part~\cite{luo2024fmb}, and screwing would require rotating the entire arm around the screw axis rather than spinning the part within the fingers~\cite{yin2026emergent}.

\begin{figure}[h!]
  \centering
  \includegraphics[width=0.99\textwidth]{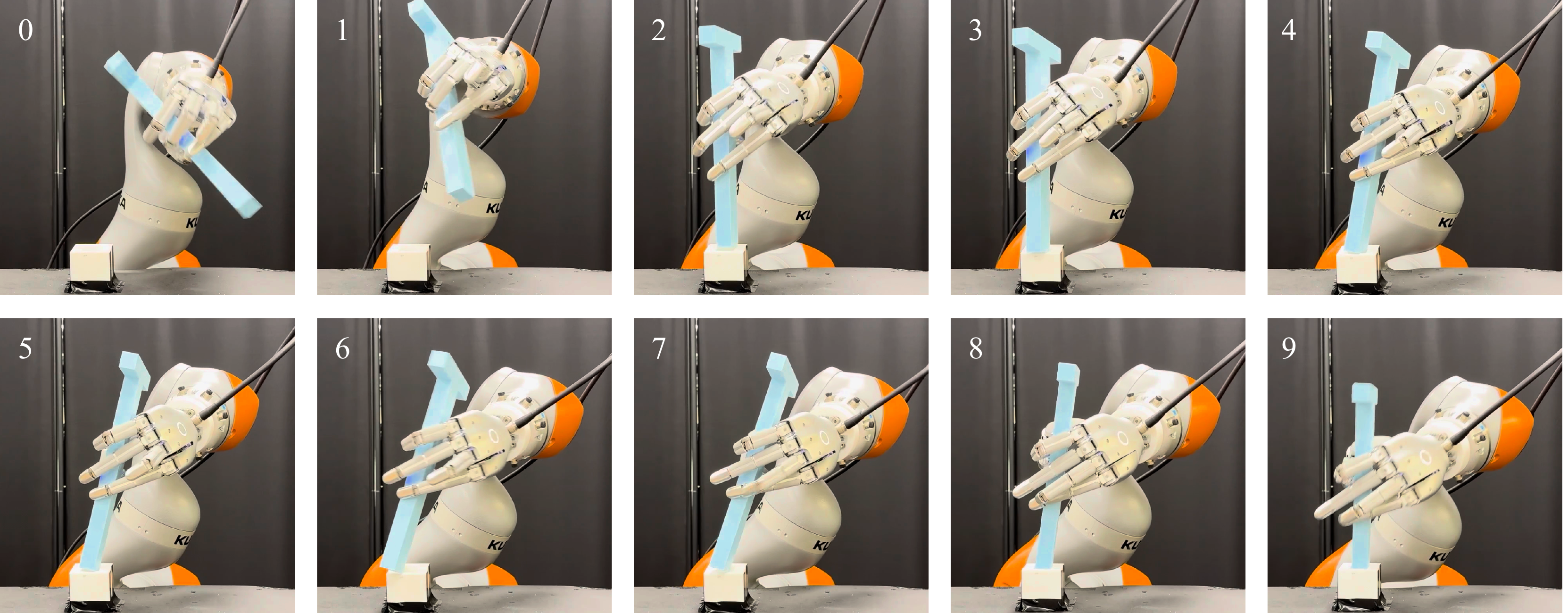}
  \caption{\textbf{Tilted Insertion and Local Search.}
  A representative real-world \texttt{Tight-Insertion} rollout. The policy approaches the hole with a tilted insertion strategy, makes contact near the fixture, and performs a local search with small corrective motions before committing to insertion.}
  \label{fig:local-search-tight-insertion}
\end{figure}

Fig.~\ref{fig:local-search-tight-insertion} shows a representative real-world \texttt{Tight-Insertion} rollout. After finetuning, the policy does not move directly to the final pose. Instead, it approaches the hole with a tilted insertion strategy, makes contact near the fixture, and uses small closed-loop corrective motions to search locally before committing to insertion. This behavior becomes especially important at tighter clearances, where small pose errors are enough to block direct insertion.

\begin{figure}[h!]
  \centering
  \includegraphics[width=\textwidth]{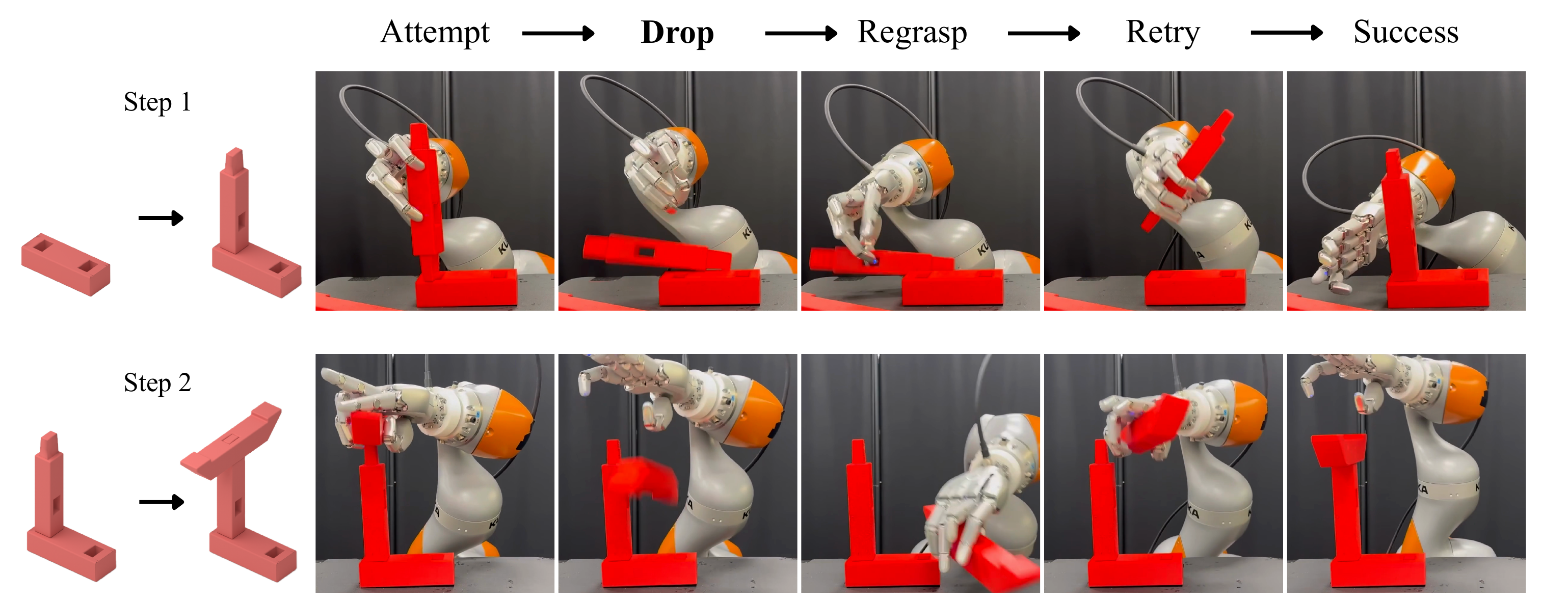}
  \caption{\textbf{Recovery After Drops.}
  Representative real-world \texttt{Assemble-Beam} rollouts for both assembly steps. After dropping the part, the policy continues acting closed-loop, regrasps the object, retries the
  assembly motion, and completes the task without a scripted recovery controller.}
  \label{fig:recovery-retry}
\end{figure}

Fig.~\ref{fig:recovery-retry} shows that the policy remains closed-loop after early failures. Even after an initial failed grasp or drop, the policy continues acting from the new state, regrasps the part, retries the assembly motion, and completes the task without a scripted recovery controller.

\textbf{Failure modes.}
Real-world failures arise from both perception and contact dynamics. Perception remains a major failure mode even on larger parts: fast part motion, hand-object occlusion, and visually similar objects can cause the pose estimator to lose track of the manipulated part. Screwing is especially challenging because the policy must track object rotation, while the rectangular leg has approximate 90$^\circ$ rotational symmetries that can confuse pose estimation and cause the policy to rotate in the wrong direction. We add colored tape to each side of the rectangular leg to make the orientations visually distinguishable, but pose estimation remains imperfect during fast rotations and occlusions. Control failures typically occur during the final contact-rich phase, when the policy repeatedly attempts insertion but fails to align with the hole, or contacts the fixture in ways that differ from simulation. In simulation, fixtures are rigid and immovable, while in the real setup, they are taped to a foam tabletop for safety and can move or comply under contact. This behavior is never observed in simulation and can cause the policy to struggle when its corrective motions no longer produce the expected relative motion between the part and fixture. We also observe some failed in-hand reorientations and drops during rotation before insertion, requiring the policy to regrasp. In contrast, grasp acquisition is highly reliable: real-world failures rarely arise from missed grasps except when perception fails catastrophically.

\section{Unified vs. Task-Specific Policies}
\begin{figure}[h]
  \centering
  \includegraphics[width=0.65\linewidth]{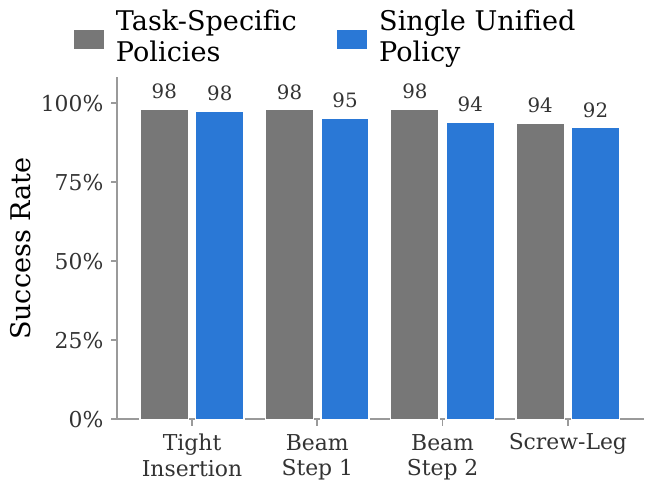}
  \caption{\textbf{Unified vs. Task-Specific Policies.} A unified policy trained on all tasks achieves comparable success rates to task-specific policies.}
  \label{fig:additional-unified}
\end{figure}

While previous experiments used \methodname{} to train single-task policies, we next evaluate whether it can be used to train a single policy that performs multiple assembly tasks. We co-train a unified policy across all four tasks. Fig.~\ref{fig:additional-unified} shows that the unified policy achieves comparable success rates to the task-specific policies.

\section{Training with Movable Fixtures}
\begin{figure}[h]
  \centering
  \includegraphics[width=0.65\linewidth]{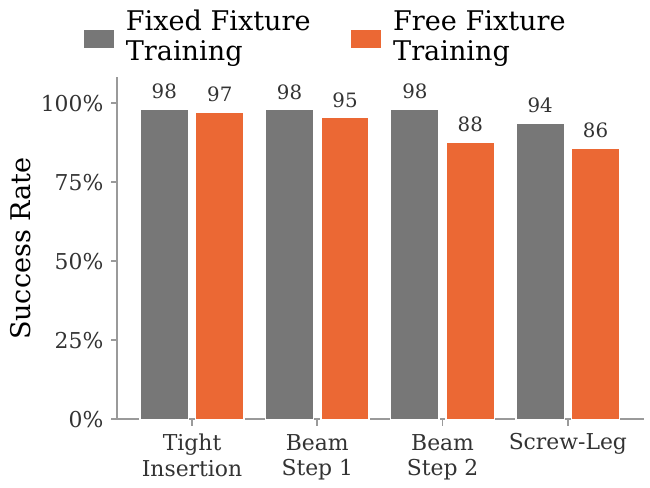}
  \caption{\textbf{Training with Movable Fixtures.} Policies can be successfully trained with free, movable fixtures. Performance is compared against policies trained with fixed fixtures.}
  \label{fig:additional-movable-fixtures}
\end{figure}

While previous experiments used rigidly fixed fixtures, we next evaluate whether \methodname{} can learn assembly with free, movable fixtures. We train and evaluate policies with these free fixtures. Fig.~\ref{fig:additional-movable-fixtures} shows that performance remains close to fixed-fixture training for \texttt{Tight-Insertion} and \texttt{Assemble-Beam} Step~1, while the fixtures in \texttt{Assemble-Beam} Step~2 and \texttt{Screw-Leg} are more easily displaced during contact.

\section{Simulation and Computational Resources}
\label{sec:simulation-compute}

All policies are trained in Isaac Sim on a single NVIDIA RTX A6000 GPU. The physics simulation runs at 120 Hz, while the policy outputs actions at 60 Hz. Play pretraining uses 24,576 parallel simulation environments and is run for 7 days. Each downstream assembly policy is finetuned for 1 day using 12,288 parallel environments. We use fewer environments during finetuning because modeling contact-rich assembly interactions requires more GPU memory than the free-space manipulation used during play pretraining. For each wall-clock comparison, all methods within the same figure are run on the same GPU type with identical hyperparameters, environment counts, and training budgets.

\section{Policy Architecture and RL Algorithm}\label{sec:policy-rl-details} We train the play policy using Split and Aggregate Policy Gradients (SAPG)~\cite{sapg2024}, a population-based variant of PPO that improves exploration in massively parallel environments. The actor uses an LSTM to integrate interaction history and infer unobserved object properties, followed by a multilayer perceptron that outputs the arm and hand actions. We use an asymmetric actor--critic architecture. While the actor receives only the observations available at deployment, the critic additionally receives noise-free and undelayed observations, palm and object velocities, reward signals, and stateful progress features such as the minimum goal distance reached and whether the object has been lifted. This privileged information is used only during training. The same RL algorithm and hyperparameters are used for both pretraining and finetuning.

\begin{table}[t] \centering \small \setlength{\tabcolsep}{6pt} \caption{\textbf{Play policy architecture and SAPG hyperparameters.}} \label{tab:play-policy-hyperparameters} \begin{tabular}{@{}ll@{}} \toprule \textbf{Parameter} & \textbf{Value} \\ \midrule Actor network & LSTM[1024] + MLP[1024, 1024, 512, 512] \\ Critic network & MLP[1024, 1024, 512, 512] \\ Learning rate & $1 \times 10^{-4}$ \\ Minibatch size & 98{,}304 \\ SAPG block size & 4{,}096 \\ Entropy bonus scale & 0.002 \\ Discount factor $\gamma$ & 0.99 \\ GAE parameter $\lambda$ & 0.95 \\ PPO clip range & 0.1 \\ \bottomrule \end{tabular} \end{table}

\section{Play Pretraining Details} \label{sec:play-pretraining-details} 

\textbf{Pretraining Environment and Procedural Objects.} At the start of each episode, a procedurally generated object is placed at a random pose on the table. The robot must grasp and lift the object, then manipulate it through a sequence of randomly sampled 6D goal poses. Each object is formed by rigidly combining two cuboid or capsule primitives. The primary component defines the graspable region, and its length and cross-sectional dimensions are sampled from $[5,30]$ cm. A secondary component is attached near one end, with length sampled from $[1,15]$ cm and cross-sectional dimensions from $[0.5,12]$ cm. We independently randomize the component densities, sampling the primary component from $[300,600]~\mathrm{kg/m^3}$ and the secondary component from $[300,2000]~\mathrm{kg/m^3}$. This produces broad variation in geometry, mass, center of mass, and inertia. 

\textbf{Episode Initialization and Goal Sampling.} At the start of each episode, the robot is initialized around a default joint configuration with uniform noise of $\pm 0.1$ rad. The object is placed above the center of the table with its position randomized within $\pm 10$ cm and its orientation sampled randomly. The first goal is sampled within the robot's reachable workspace, with position relative to the table center sampled as $x \in [-0.35,0.35]$ m, $y \in [-0.1,0.2]$ m, and $z \in [0.15,0.52]$ m. After reaching a goal, the next goal is sampled relative to it with a translation of up to $0.1$ m and a rotation of up to $90^\circ$. This produces smooth but diverse goal trajectories that require repeated in-hand reorientation. Each episode lasts at most 600 control steps (10 s). It terminates early if the object falls below the table, the hand moves more than $1.5$ m from the object, the contact force measured at the table exceeds $100$ N, or the maximum number of goal successes is reached. 

\textbf{Keypoint-Based Pose Representation.} We represent each 6D object pose using four keypoints defined in the local object frame. Given dimensions $\mathbf{s}=[s_x,s_y,s_z]$, the keypoints are placed at \begin{equation} \mathcal{K}(\mathbf{s}) = \left\{ \begin{bmatrix} s_x/2 \\ s_y/2 \\ s_z/2 \end{bmatrix}, \begin{bmatrix} s_x/2 \\ -s_y/2 \\ -s_z/2 \end{bmatrix}, \begin{bmatrix} -s_x/2 \\ s_y/2 \\ -s_z/2 \end{bmatrix}, \begin{bmatrix} -s_x/2 \\ -s_y/2 \\ s_z/2 \end{bmatrix} \right\}. \end{equation} For a pose $o=(R_o,\mathbf{t}_o)$, each keypoint is transformed into the world frame as \begin{equation} \mathbf{o}_i = R_o\mathbf{k}_i+\mathbf{t}_o. \end{equation} We define the distance between current and goal poses as \begin{equation} d(o,g)=\max_i \left\|\mathbf{o}_i-\mathbf{g}_i\right\|_2. \end{equation} This provides a single metric that jointly captures translation and rotation error. For policy observations, we define the keypoints using the dimensions of the object's primary component. For reward computation, we instead use fixed dimensions $\mathbf{s}^{\mathrm{rew}}=[0.14,0.03,0.03]$ m, ensuring a consistent trade-off between translational and rotational errors across objects. 

\textbf{Policy Observations and Action Space.} The policy receives a 140-dimensional observation containing robot proprioception, the current object pose, the goal pose, and an object geometry descriptor. Robot proprioception includes the 29 joint positions and velocities, previous joint-position targets, palm pose, and the positions of the five fingertips relative to the palm. We represent the current and goal object poses using four keypoints defined by the dimensions of the object's primary component. The policy observes the object orientation, the current keypoints relative to the palm, the keypoint displacements from the current pose to the goal, and the three object dimensions. Using relative keypoint representations reduces dependence on absolute workspace coordinates while conditioning the policy on object geometry. The policy outputs 29 joint-position commands for the 7-DoF arm and 22-DoF hand. Arm actions are represented as delta joint-position targets, while hand actions are represented as absolute joint-position targets. All commands are clipped to the corresponding joint limits and smoothed using an exponential moving average with coefficient $\alpha=0.1$. 

\textbf{Play Reward and Success Criterion.} The play reward combines smoothness, grasping, and goal-reaching terms: \begin{equation} r = r_{\mathrm{smooth}} + r_{\mathrm{grasp}} + \mathbb{I}_{\mathrm{grasped}}r_{\mathrm{goal}}. \end{equation} We encourage smooth control by penalizing arm and hand joint velocities: \begin{equation} r_{\mathrm{smooth}} = -\lambda_{\mathrm{arm}} \left\|\dot{\mathbf{q}}^{\mathrm{arm}}\right\|_1 -\lambda_{\mathrm{hand}} \left\|\dot{\mathbf{q}}^{\mathrm{hand}}\right\|_1. \end{equation} The grasp reward first encourages the fingertips to approach the object and then lift it from the table: \begin{align} r_{\mathrm{grasp}} &= r_{\mathrm{approach}} +(1-\mathbb{I}_{\mathrm{grasped}})r_{\mathrm{lift}}, \\ r_{\mathrm{approach}} &= \lambda_{\mathrm{approach}} \max\!\left(\bar{d}^{*}_{\mathrm{ft}}-\bar{d}_{\mathrm{ft}},0\right), \\ r_{\mathrm{lift}} &= \lambda_{\mathrm{lift}} \max(z-z_{\mathrm{init}},0) +B_{\mathrm{lifted}} \mathbb{I}[z\geq z_{\mathrm{lifted}}], \end{align} where $\bar{d}_{\mathrm{ft}}$ is the mean fingertip-to-object distance and $\bar{d}^{*}_{\mathrm{ft}}$ is the smallest distance reached so far in the episode. We set $\mathbb{I}_{\mathrm{grasped}}=1$ once the object has been lifted by 10 cm. After the object is grasped, the policy is rewarded for making progress toward the current 6D goal: \begin{equation} r_{\mathrm{goal}} = \lambda_{\mathrm{goal}} \max\!\left(d^{*}-d(o_t,g_t),0\right) +B_{\mathrm{succ}} \mathbb{I}[d(o_t,g_t)<\epsilon], \end{equation} where $d^{*}$ is the smallest goal distance reached since the current goal was sampled. Reaching a goal within $\epsilon=1$ cm yields a sparse success bonus, after which a new goal is sampled. For reward computation, we use the keypoint-based pose distance defined above with fixed dimensions $\mathbf{s}^{\mathrm{rew}}=[0.14,0.03,0.03]$ m. This provides a consistent trade-off between translation and rotation error across objects. We use $\lambda_{\mathrm{arm}}=0.03$, $\lambda_{\mathrm{hand}}=0.003$, $\lambda_{\mathrm{approach}}=50$, $\lambda_{\mathrm{lift}}=20$, $\lambda_{\mathrm{goal}}=200$, $B_{\mathrm{lifted}}=300$, and $B_{\mathrm{succ}}=1000$. \begin{table}[t] \centering \small \setlength{\tabcolsep}{6pt} \caption{\textbf{Domain randomization used during play pretraining.}} \label{tab:play-domain-randomization} \begin{tabular}{@{}ll@{}} \toprule \textbf{Parameter} & \textbf{Randomization} \\ \midrule Current/goal pose noise (translation) & $1$ cm \\ Current/goal pose noise (rotation) & $5^\circ$ \\ Object-pose delay & $0$--$10$ control steps \\ Action and proprioception delay & $0$--$3$ control steps \\ Joint-velocity observation noise & $\sigma=0.1$ rad/s \\ Object-dimension scale & $90\%$--$110\%$ \\ Table-height variation & $\pm 1$ cm \\ External force perturbation & $20.0$ N \\ External torque perturbation & $2.0$ N\,m \\ \bottomrule \end{tabular} \end{table} 

\textbf{Domain Randomization.} We apply domain randomization to observations, action execution, object geometry, and environment dynamics to improve sim-to-real transfer. We perturb the observed current and goal object poses, introduce latency in object-pose estimates, actions, and proprioceptive observations, and add noise to joint-velocity observations. We add noise to the object geometry descriptor and vary table height, and apply random forces and torques to the manipulated object, encouraging the policy to learn stable grasps and recovery behaviors. Table~\ref{tab:play-domain-randomization} lists the values.

\section{Pretraining Ablation Implementations} \label{sec:pretraining-ablation-implementations} We ablate four components of play pretraining: object diversity, training objective, trajectory diversity, and goal precision. Unless otherwise stated, all variants use the same environment, policy architecture, reward coefficients, domain randomization, training budget, and downstream assembly-finetuning procedure as the default \methodname{} policy. Each ablation changes only the pretraining factor under study. 

\textbf{Object Diversity.} We vary the number of procedurally generated objects used during pretraining among 10, 100, and 1000, with 1000 objects used by default. Each object set is sampled from the same procedural distribution described in Sec.~\ref{sec:play-pretraining-details}, including the same geometry and density ranges. During pretraining, each environment samples an object from the corresponding fixed object set. All goal-sampling and optimization settings remain unchanged. 

\textbf{Training Objective.} We compare the default 6D pose-reaching objective against Translation-only and Rotation-only variants. The default objective requires jointly matching the target translation and orientation using the keypoint-based pose distance. Translation-only uses the same sampled goal sequences, but computes goal success using only the Euclidean distance between the current and target object positions. The target orientation is ignored. Rotation-only first requires the policy to grasp, lift, and move the object to a randomly sampled 6D pose. Subsequent goals keep this target position fixed and vary only the target orientation, requiring repeated in-hand reorientation without translational goal changes.

\textbf{Trajectory Diversity.} The default policy samples every goal trajectory online during pretraining. To study the effect of trajectory diversity, we instead construct fixed banks containing either 10 or 100 goal trajectories. Each trajectory follows the same sampling distribution as the default setting, including the same workspace bounds and maximum translation and rotation between consecutive goals. During pretraining, episodes repeatedly sample from the corresponding fixed trajectory bank rather than generating new trajectories online. 

\textbf{Goal Precision.} We vary the success tolerance used to determine whether a play goal has been reached. The default policy uses $\epsilon=1$ cm, while the ablations use $\epsilon=5$ cm or $\epsilon=10$ cm. Only the success threshold is changed. The sampled goals, pose representation, reward coefficients, and all other training settings remain identical.
\section{Assembly Finetuning Details}\label{sec:assembly-finetuning-details}

\textbf{CAD-to-RL Automation.} The CAD model provides geometry and goal poses in the assembled state. Our pipeline automatically derives insertion axes and pre-contact poses from the geometry. For example, for the screw task, the thread axis, pitch, and depth geometry define goals at 90$^\circ$ increments. Mating regions in the CAD file are automatically identified and are modeled with SDF meshes, while other parts of the CAD use standard convex decomposition. Grasp bounding boxes use the object bounds, and all tasks use the same fixed 1cm goal tolerance without task-specific tuning.

\textbf{Environment Construction from CAD.} For each assembly step, we import the manipulated part and fixture CAD models into Isaac Sim as rigid bodies. Most geometry is represented using convex decomposition for efficient simulation. However, convex approximations can distort narrow holes and mating interfaces, changing the effective clearance and contact dynamics. We therefore represent only the contact-critical hole and insertion components using signed distance fields (SDFs) at resolution $256$. This targeted hybrid representation provides detailed collision geometry where precision matters while avoiding the memory and computational cost of applying high-resolution SDFs to the entire assembly in simulation.

\textbf{Episode Initialization and Reset Distribution.} At the beginning of each episode, the manipulated part is spawned above the table with its planar position sampled independently as $x,y \in [-0.1,0.1]$ m and its orientation sampled uniformly at random. The part is then dropped onto the table, producing diverse stable initial poses. The fixture is placed flat at a random location on the table. 

\textbf{CAD-Derived Assembly Goal Sequences.} For each assembly step, the CAD design specifies the desired transform $T^{f}_{p}$ of the manipulated part $p$ in the fixture frame $f$. Given the current fixture pose $f_t \in SE(3)$, we compute the final world-frame goal as \begin{equation} g_M = f_t T^{f}_{p}. \end{equation} This makes the goal invariant to the randomized fixture placement. For contact-rich tasks, we augment the final assembled pose with a small sequence of sparse intermediate goals obtained by reversing the corresponding disassembly motion. For insertion tasks, this includes an aligned pre-insertion pose immediately before contact, followed by the final inserted pose. For screwing, we generate successive goals along the thread with $90^\circ$ rotational offsets. The policy advances to the next goal once the current pose is reached within the specified tolerance, and the final goal corresponds to successful completion of the assembly step.

\textbf{Assembly Reward, Goal Progression, and Success.} Assembly finetuning uses a sparse task reward derived only from the CAD-defined goal sequence. We remove the grasping, lifting, and dense pose-progress rewards used during play pretraining. The finetuning reward contains only a smoothness regularizer and sparse bonuses for reaching the assembly goals: \begin{equation} r_t = r_{\mathrm{smooth}} + r_{\mathrm{goal}}. \end{equation} We retain the same smoothness regularizer as during pretraining: \begin{equation} r_{\mathrm{smooth}} = -\lambda_{\mathrm{arm}} \left\|\dot{\mathbf{q}}^{\mathrm{arm}}\right\|_1 -\lambda_{\mathrm{hand}} \left\|\dot{\mathbf{q}}^{\mathrm{hand}}\right\|_1. \end{equation} Let $g_m$ denote the active CAD-derived goal. The policy receives no task-dependent reward as it approaches the goal. It receives a sparse success bonus only once the part enters the goal tolerance: \begin{equation} r_{\mathrm{goal}} = B_{\mathrm{succ}} \mathbb{I}\!\left[d(o_t,g_m)<\epsilon\right] + r_{\mathrm{retract}}, \end{equation} where $d(\cdot,\cdot)$ is the keypoint-based 6D pose distance defined in Sec.~\ref{sec:play-pretraining-details}. When an intermediate goal is reached within $\epsilon=1$ cm, the environment advances to the next goal in the sequence. Reaching the final goal marks the assembly itself as successful. For the final goal, we additionally provide a sparse retraction bonus once the assembled part remains at its goal and the robot moves its palm more than $0.2$ m away: \begin{equation} r_{\mathrm{retract}} = B_{\mathrm{retract}} \mathbb{I}\!\left[ d(o_t,g_M)<\epsilon \;\land\; \left\|\mathbf{p}^{\mathrm{palm}}_t -\mathbf{p}^{\mathrm{obj}}_t\right\|_2 > 0.2~\mathrm{m} \right]. \end{equation} This encourages the robot to release and retract after completing the assembly, rather than relying on continued hand contact to hold the part at its final pose. We use $B_{\mathrm{succ}}=B_{\mathrm{retract}}=1000$. Thus, apart from action smoothness, the environment provides feedback only when the policy reaches a CAD-derived goal or completes the final retraction. There are no explicit rewards for approaching, grasping, lifting, alignment, contact, or reducing pose error. These behaviors must instead be retained from the pretrained play policy and adapted through sparse-reward finetuning.

\textbf{Domain Randomization During Finetuning.} We apply domain randomization during assembly finetuning to improve robustness to perception errors, control latency, and contact-dynamics mismatch. We perturb the observed current part pose and CAD-derived goal pose with independent translational and rotational noise. Goal-pose noise models errors in the estimated fixture pose used to compute the assembly goals. We also randomize the physical fixture yaw, object-pose latency, action and proprioception latency, joint-velocity observations, observed part dimensions, and table height. Finally, random external forces and torques are applied to the manipulated part to encourage stable grasping and recovery during contact-rich interactions. Table~\ref{tab:finetuning-domain-randomization} summarizes the parameters. 

\begin{table}[h] \centering \small \setlength{\tabcolsep}{6pt} \caption{\textbf{Domain randomization used during assembly finetuning.}} \label{tab:finetuning-domain-randomization} \begin{tabular}{@{}ll@{}} \toprule \textbf{Parameter} & \textbf{Randomization} \\ \midrule Current part-pose noise (translation) & $1$ cm \\ Current part-pose noise (rotation) & $5^\circ$ \\ Goal-pose noise (translation) & $2$ mm \\ Goal-pose noise (rotation) & $1^\circ$ \\ Fixture yaw & $[-10^\circ,10^\circ]$ \\ Object-pose delay & $0$--$10$ control steps \\ Action and proprioception delay & $0$--$3$ control steps \\ Joint-velocity observation noise & $\sigma=0.1$ rad/s \\ Part-dimension noise scale & $90\%$--$110\%$ \\ Table-height variation & $\pm 1$ cm \\ External force perturbation & $20.0$ N \\ External torque perturbation & $2.0$ N\,m \\ \bottomrule \end{tabular} \end{table}

\raggedbottom
\section{Inference Time Pipeline}
\label{app:inference}

\begin{figure}[h!]
  \centering
  \includegraphics[width=\textwidth]{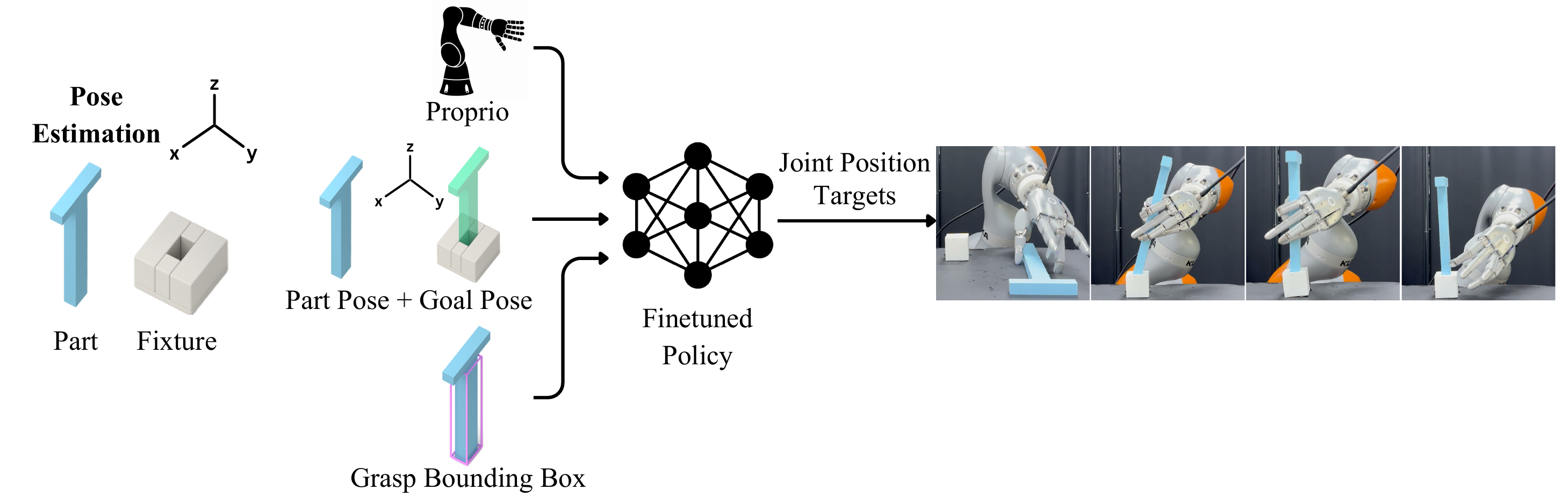}
  \caption{
  \textbf{Inference-Time Pipeline.}
    At deployment, CAD models are reused to estimate the current part pose, compute CAD-derived goal poses relative to the fixture, and define the grasp bounding box. The finetuned policy takes robot proprioception, part pose, goal pose, and the grasp bounding box as input, and outputs joint position targets for the arm and dexterous hand.
  }
  \label{fig:inference_time_pipeline}
  \vspace{3mm}
\end{figure}

At deployment, we reuse the part and fixture CAD models from training in three ways, as illustrated in Fig.~\ref{fig:inference_time_pipeline}. (1) Part pose estimation: FoundationPose~\cite{wen2024foundationpose} tracks the manipulated part pose online and provides the current object pose to the policy. (2) Goal pose: the fixture pose is estimated once at the beginning of each rollout. Since the fixture remains fixed during execution, the desired part goal poses are then computed from the estimated fixture pose and the CAD-specified assembly transforms. (3) Grasp bounding box: we define a grasp bounding box on the part CAD model that specifies the region where the robot should grasp the object.

The finetuned policy runs as a closed-loop controller at 60 Hz. At each control step, it receives robot proprioception, the current part pose, the desired goal pose, and the grasp bounding box. The policy outputs joint position targets for the 7-DoF arm and 22-DoF dexterous hand, which are executed by the robot's low-level joint position controllers. The part pose tracker runs at 30 Hz, and the controller reuses the most recent pose estimate between tracking updates. We do not use an additional scripted insertion, screwing, or recovery controller at deployment. Contact-rich behaviors such as local search, corrective motions, regrasping, and in-hand spinning are produced by the learned policy.

For each assembly task, the policy follows a short sequence of sparse goal poses $\mathcal{G}=\{g_1,\ldots,g_M\}$ derived from the CAD assembly. At the beginning of a rollout, the active goal is initialized to the first goal $g_1$. During execution, we compute the pose distance between the current part pose $o_t$ and the active goal $g_m$. When $d_{\mathrm{pose}}(o_t,g_m) < \epsilon_{\mathrm{goal}}$, the controller advances to the next goal, $g_{m+1}$. This process repeats until the final goal $g_M$ is reached, at which point the assembly is considered complete. For example, insertion tasks use intermediate pre-insertion goals before the final inserted pose, while screwing tasks use goal poses along the thread to induce axial rotation.

\begin{figure}[H]
  \centering
  \includegraphics[width=\textwidth]{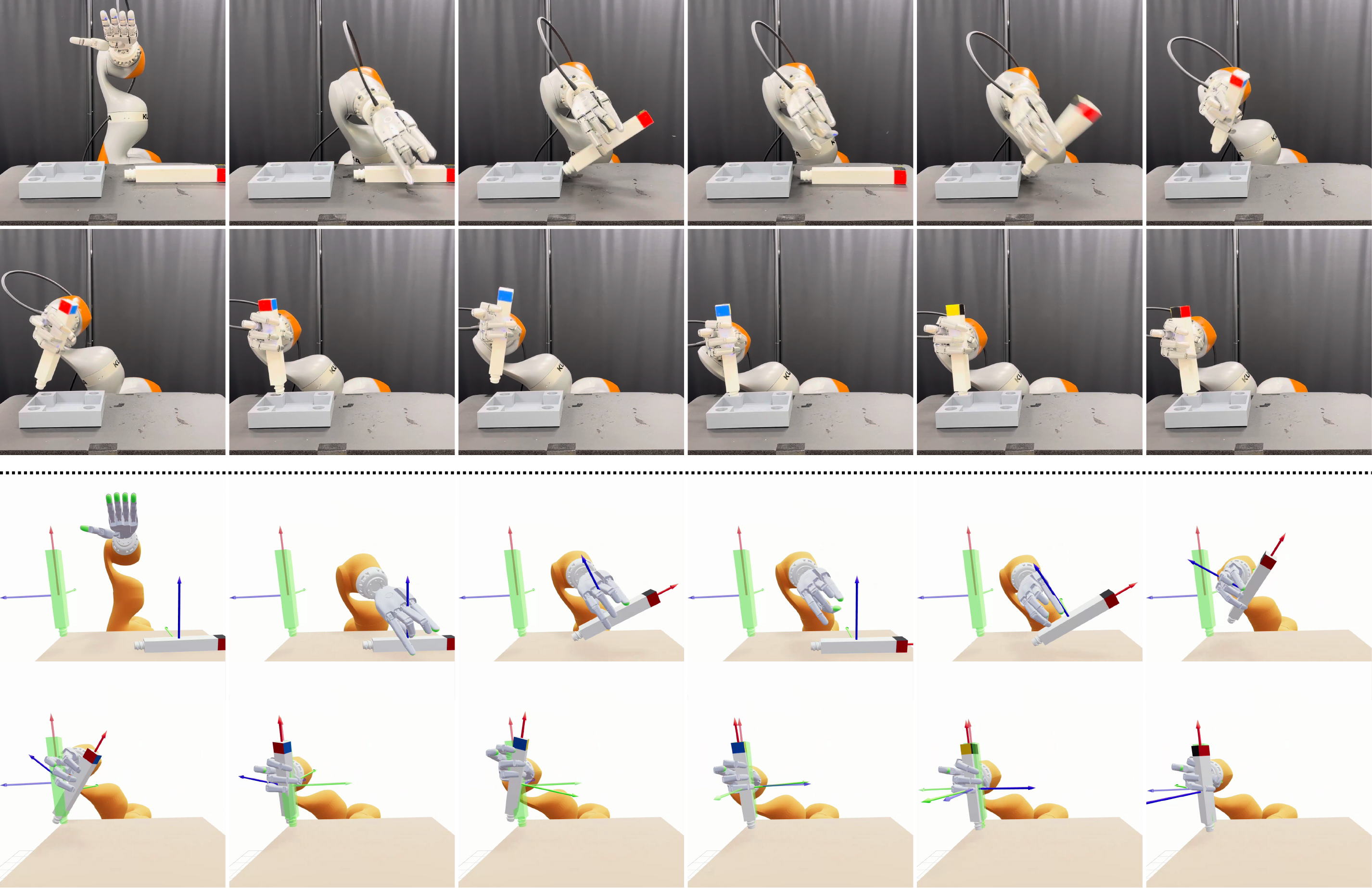}
  \caption{\textbf{Policy Observations During Real-World Deployment.}
  (Top) A representative real-world \texttt{Screw-Leg} rollout, with (Bottom) the corresponding policy observations. The translucent green object and translucent axes
  denote the active sparse goal pose, while the opaque object and full-opacity axes denote the estimated current part pose observed by the policy. As the rollout progresses, the active
  goal advances along the screw axis with successive rotational offsets, and the policy tracks these goals by reorienting the leg in-hand. During fast motions, the visualized current pose can lag the real object due to object-pose tracking latency.}
  \label{fig:policy-observation-rollout}
\end{figure}

Fig.~\ref{fig:policy-observation-rollout} visualizes the policy observations during a real-world \texttt{Screw-Leg} rollout. The policy
receives the estimated current part pose and the active sparse goal pose at each timestep. As the rollout progresses, the active goal advances through the CAD-derived goal sequence.

\clearpage

\end{document}